\documentclass[lettersize,journal]{IEEEtran}

\usepackage{times}

\usepackage[numbers]{natbib}
\usepackage{multicol}
\usepackage[linkcolor=blue,colorlinks=true, bookmarks=true]{hyperref}

\usepackage{graphicx}
\usepackage{amsmath}%,amsfonts}
\usepackage{amssymb}

\usepackage{amsthm}
\usepackage{mathtools}
\usepackage{cuted}
\usepackage{soul}
\usepackage[dvipsnames]{xcolor}

\def\BibTeX{{\rm B\kern-.05em{\sc i\kern-.025em b}\kern-.08em
    T\kern-.1667em\lower.7ex\hbox{E}\kern-.125emX}}
\usepackage{balance}

\newtheorem{theorem}{Theorem}

\newtheorem{remark}{Remark}

\usepackage{subcaption}
\usepackage{caption}
\DeclareCaptionType{equ}[][]

\usepackage{comment}
\usepackage{algorithm}  
\usepackage{algpseudocode}[noend]

\graphicspath{{figures/}}

\newcommand{\reals}{\mathbb{R}}
\newcommand{\s}{\mathcal{S}}
\newcommand{\X}{\mathcal{X}}
\newcommand{\U}{\mathcal{U}}
\newcommand{\C}{\mathcal{C}}

\newcommand{\eqn}[1]{\begin{align}#1\end{align}}
\newcommand{\Ep}{\mathbb{E} }

\DeclareMathOperator*{\argmax}{arg\,max}
\DeclareMathOperator*{\argmin}{arg\,min}

\newcommand{\blue}[1]{{\normalsize{{\color{blue}#1}}}}
\newcommand{\probs}{\mathbb P}

\begin{document}

\title{FORESEE: Prediction with Expansion-Compression Unscented Transform for Online Policy Optimization}
%A Finite-Particle Perspective on State Prediction for Online Policy Optimization}

\author{Hardik Parwana, \IEEEmembership{Member, IEEE} Dimitra Panagou, \IEEEmembership{Senior Member, IEEE}
\thanks{Hardik Parwana and Dimitra Panagou are with the Department of Robotics, University of Michigan, MI 48109, USA (email: hardiksp@umich.edu, dpanagou@umich.edu). }
        % <-this % stops a space
\thanks{This work was partially sponsored by the Office of Naval Research (ONR), under grant number N00014-20-1-2395. The views and conclusions contained herein are those of the authors only and should not be interpreted as representing those of ONR, the U.S. Navy or the U.S. Government.}
}

% The paper headers
\markboth{Journal of \LaTeX\ Class Files,~Vol.~14, No.~8, August~2021}%
{Shell \MakeLowercase{\textit{et al.}}: A Sample Article Using IEEEtran.cls for IEEE Journals}

%\IEEEpubid{0000--0000/00\$00.00~\copyright~2021 IEEE}
% Remember, if you use this you must call \IEEEpubidadjcol in the second
% column for its text to clear the IEEEpubid mark.

\maketitle

\begin{abstract}
Propagating state distributions through a generic, uncertain nonlinear dynamical model is known to be intractable and usually begets numerical or analytical approximations. We introduce a method for state prediction, called the Expansion-Compression Unscented Transform, and use it to solve a class of online policy optimization problems. Our proposed algorithm propagates a finite number of sigma points through a state-dependent distribution, which dictates an increase in the number of sigma points at each time step to represent the resulting distribution; this is what we call the expansion operation. To keep the algorithm scalable, we augment the expansion operation with a compression operation based on moment matching, thereby keeping the number of sigma points constant across predictions over multiple time steps. Its performance is empirically shown to be comparable to Monte Carlo but at a much lower computational cost. Under state and control input constraints, the state prediction is subsequently used in tandem with a proposed variant of constrained gradient-descent for \textit{online} update of policy parameters in a receding horizon fashion. The framework is implemented as a differentiable computational graph for policy training. We showcase our framework for a quadrotor stabilization task as part of a benchmark comparison in safe-control-gym and for optimizing the parameters of a Control Barrier Function based controller in a leader-follower problem.

\end{abstract}


\section{Introduction}
Achieving robot autonomy is particularly challenging in uncertain environments due to the risk of failures and fatal crashes. The ability to learn and adapt the parameters of the control policy by quantifying the uncertainties present in the system, and evaluating the performance of the policy as a function of its parameters, is crucial to enhancing robot autonomy. We introduce a new model-based Reinforcement Learning (RL) algorithm FORESEE (4C): Foresee for Certified Constrained Control, that uses stochastic estimates of system dynamics to predict the future state distribution, and then adapts policy parameters to improve performance while ensuring the satisfaction of state and control input constraints. To this end, in this paper we consider the problem of optimizing the parameters of a model-based Reinforcement Learning policy under uncertain states. We propose a novel, computationally-efficient and scalable method for the online prediction of nonlinear state distributions using the Unscented Transform, and interface it with a constraint-aware gradient-descent policy that adapts the policy parameters online for performance improvement.

Performance, often expressed as a cumulative cost induced over a time horizon, is usually optimized for by solving a stochastic optimization problem over policy parameters. RL and Differential Dynamic Programming (DDP) are two frameworks that have built upon the success of gradient-based methods to incrementally improve policies over time. Our focus in this paper is on the model-based variants of these frameworks, which have shown to be efficient by taking into account the prior knowledge about the system, compared to model-free variants that usually require prohibitive amount of training data \cite{deisenroth2013survey}, and therefore are not suitable for real-time training and adaptation in novel environments. 

\subsection{Earlier work on state prediction for nonlinear dynamics}

Performance evaluation typically involves predicting the distribution of the reward over a time horizon by propagating the state over stochastic system dynamics over multiple time steps. This is, in general, analytically intractable for nonlinear systems \cite{deisenroth2013gaussian}. Several approximate analytical and numerical methods have therefore been proposed. A popular one that is still prevalent was proposed in PILCO \cite{deisenroth2011pilco} and starts with a Gaussian state distribution, passes it through a nonlinear transformation model, approximates the resulting non-Gaussian distribution with a Gaussian distribution, and then repeats this procedure. When the transition dynamics is expressed as a Gaussian Process (GP) with Gaussian kernel, PILCO \cite{deisenroth2011pilco} derived analytical formulas for propagating the exact mean and covariance of the system state across a time step, thereby converting the stochastic optimization problem into a deterministic one.
%with well-defined dynamics for mean and covariance of the system states.
The analytical gradients were then used to update the policy parameters with gradient descent. 

The idea since then has also been used in several other works on stochastic Model Predictive Control (MPC) \cite{kamthe2018data}, and DDP \cite{theodorou2010stochastic, pan2014probabilistic} approaches. This method however ignores higher-order moments of the distribution, which can be significant when the original distributions have high variance, and when the transition dynamics is highly nonlinear. Furthermore, repeated approximation over multiple time steps causes the prediction to drift away from the true distribution \cite{vinogradska2018numerical}. Finally, since these analytical expressions hold only for Gaussian kernels in GP, they are restrictive in practice as different kernels are preferred depending on the environment \cite{schulz2018tutorial}. Model learning algorithms other than GP have also been used. Deep PILCO \cite{gal2016improving} uses Bayesian Neural Network (NN), but it still relies on the same mean and covariance matching principle as PILCO.
\begin{strip}
    \centering
    \includegraphics[width=0.99\textwidth]{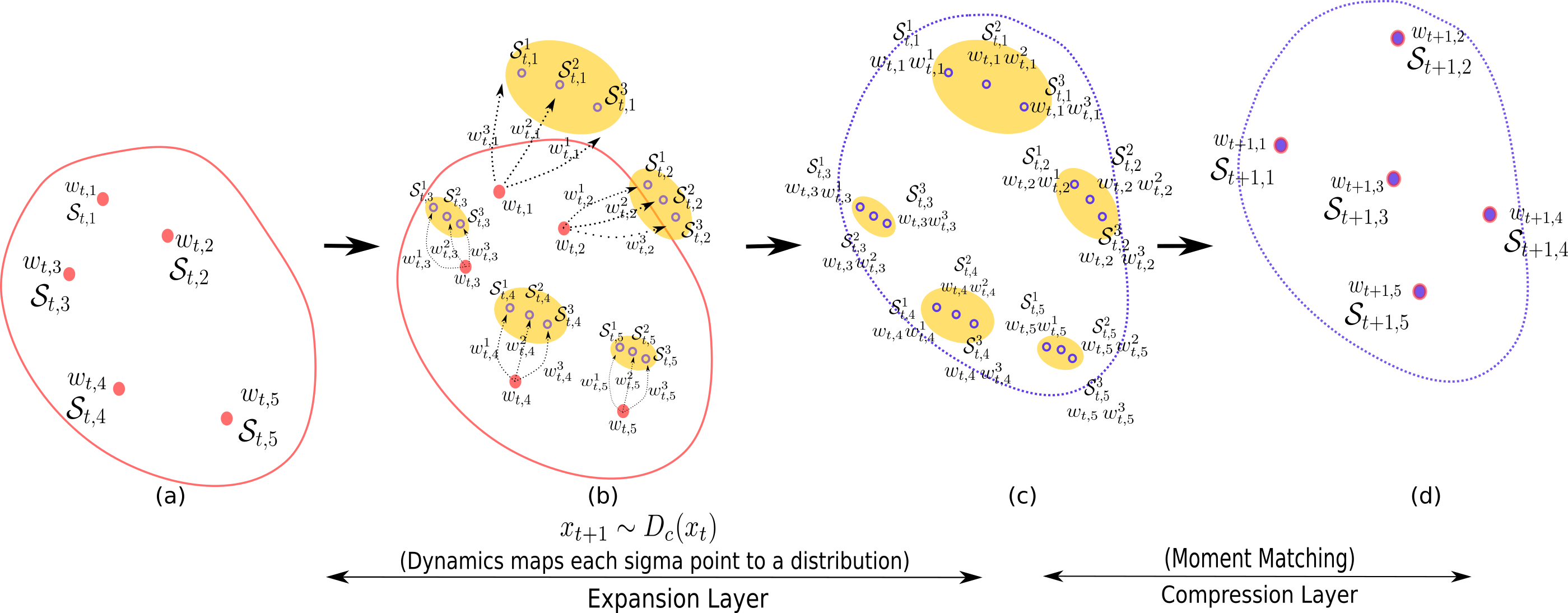}
    \captionof{figure}{\small{(a) The red particle (sigma point) $ \s_{t,i}$ and their weights $w_{t,i}$ represents the distribution of state $\hat x_t$. (b) Since the uncertainty is state-dependent, each red point $\s_{t, i}$ gives rise to the states in the yellow region, each of which is represented by three blue points $\s_{t, i}^j$ and weights $w_{t,i}^j$ in the expansion layer. (c) The overall weight of $\s_{t,i}^j$ is obtained by multiplying the weight of the edge with the weight of the corresponding root node. (d)
    %The size of the yellow region changes as the uncertainty in dynamics is state-dependent. 
    The 15 points resulting from the expansion layer are then compressed back to five red-blue particles in the compression layer to represent the blue distribution of $x_{t+1}$.}}%
    \label{fig::uncertainty_prop}%
\end{strip} 
Some works have tried to alleviate this problem by maintaining higher-order moments, or by approximating distributions with more generic parametric representations. \cite{stantonmodel} and \cite{amadio2022model} employ Monte Carlo (MC) to estimate future rewards and their gradients. \cite{ozaki2018stochastic} used Unscented Transform (UT) to propagate uncertainty in the presence of Gaussian noise, and improve an initially given feasible trajectory with DDP. \cite{vinogradska2018numerical} represented distributions as a mixture of Gaussians, maintaining coefficients for basis functions, and used Gaussian quadrature rule to propagate the coefficients over multiple time steps. 

The above methods, while leading to more accurate results than the Gaussian approximation, have their own disadvantages. MC is known to require an excessively large number of samples and is computationally expensive \cite{o1987monte, ghavamzadeh2006bayesian}. \cite{ozaki2018stochastic} used the vanilla Unscented Transform, which is known to ignore moments higher than the covariance when choosing sigma points. \cite{ozaki2018stochastic} also only considers random perturbations that are independent of the state, whereas we are interested in perturbations that are state-dependent, such as GPs. As such it can simply propagate each sigma point to a unique new sigma point and does not run into the scalability issue associated with an increasing number of sigma points in presence of state-dependent perturbations that we discuss later. \cite{vinogradska2018numerical} requires an offline analysis to design the quadrature rule. 

\subsection{Earlier gradient-based approaches for parameter update}

A key element in most of the aforementioned works is the use of gradient-based optimization for policy update. There are other approaches \cite{chatzilygeroudis2017black, levine2014learning, gal2016improving} that perform gradient-free optimization. While most of these gradient-free methods are also able to incorporate models for the system dynamics other than GPs, they either rely heavily on parallelization for efficient implementations \cite{chatzilygeroudis2017black, levine2014learning}, or require extensive offline optimization \cite{gal2016improving}. Although these methods present many advantages, our focus in this paper is on gradient-based methods that are amenable to online implementation on less powerful (computationally) platforms, in order to increase the autonomy of robotic systems.

 % in  and an ensemble of bootstrapped NN models., and \cite{gal2016improving} is a gradient-free offline optimization method. in  and an ensemble of bootstrapped NN models.

Furthermore, the learned policy is often required to satisfy strict state and input constraints. Augmenting the cost function with a penalizing term for constraint violation provides limited opportunity for formal guarantees, as the closed-loop behavior may not satisfy the hard constraints. \cite{zhang2021model} instead adds a safety layer based on control barrier functions, and \cite{cheng2019end} poses safe RL as a stochastic MPC problem. Some works directly solve the Constrained Markov Decision Problem \cite{sun2021fisar} to learn a safety-critical policy. In general, model predictive frameworks can be used in tandem with nonlinear optimization techniques to find a constraint respecting policy. For example, SafePILCO \cite{roberts2020safepilco} extends PILCO to satisfy state constraints. However, few techniques are suitable for online implementation. The approaches in \cite{andersson2015model, van2020online} use DDP to incrementally improve the trajectory online and solve the constrained RL problem. However, they only consider the mean of learned model dynamics when predicting the future. 
%In contrast to DDP approaches Unlike \cite{andersson2015model}\cite{van2020online}\cite{ozaki2018stochastic} that iteratively update a given sequence of states and control inputs, we use a parameterized controller and consider the sensitivity of current trajectory with respect to the control parameters on the

\subsection{Contributions and comparison to the authors' earlier work}

The contributions of this paper are twofold. First, we propose a novel method for predicting future state distributions for nonlinear dynamical systems with state-dependent uncertainty %with any stochastic dynamics model (i.e., beyond the standard Gaussian Process with the Gaussian kernel model). We use 
using the Unscented Transform (UT) and its variants \cite{ebeigbe2021generalized}. %to propagate distributions through nonlinear transforms given by system dynamics models with state-dependent perturbation. 
UT methods maintain a finite number of weighted \textit{particles} or \textit{sigma points} that have the same sample moments as the true distribution. \emph{However, propagating a finite number of sigma points through a model with state-dependent uncertainty requires an increase in the number of points that need to be maintained with each time step to keep track of the evolving state distribution. Therefore, in order to keep our algorithm scalable, we add a moment-preserving sigma-point contraction layer that helps keep the number of sigma points constant across each time step.} An illustration is shown in Fig.\ref{fig::uncertainty_prop}. Note that previous works do not encounter into this issue of scalability because they either considered UT with state-independent perturbation in dynamics \cite{ozaki2018stochastic}, or they performed Monte Carlo \cite{stantonmodel, amadio2022model} where owing to a large number of particles that are already employed, sampling only one state from future distribution of each particle suffices in theory in the limit of the number of particles going to infinity. 
% and hence our algorithm that are used GenUT allows us to maintain higher order moments depending on the number of sigma points employed. 
%In the simplest case, they guarantee maintenance of only the mean and covariance with Gaussian distributions, but otherwise can seamlessly correspond to the ideal Monte Carlo estimate when sigma points approach infinity. Unscented Transform, owing to the principled way in which it selects representative sigma points, is known to give much better accuracy of prediction than Monte Carlo for the same number of particles employed. 
%and hence even a few sigma points are capable to giving much more accurate results than both gaussian approximations and monte carlo. 
%Propagating uncertainty with sigma points allows us to consider more generic distributions than the Gaussian, and 
%PROF_COMMENT
\textit{To the best of our knowledge, a principled analysis to propagate a given finite number of particles through a dynamics model with state-dependent perturbation distribution is missing, and we address it in this paper.} Our empirical results show that with a finite number of sigma points\footnote{Note that in theory, one would need an infinite number of sigma points to represent all infinite moments of any arbitrary distribution, however that would be intractable in practice. In general, how to find sigma points is still an active area of research.}, our method has satisfactory performance, comparable to Monte Carlo simulations but at a much lower computational cost.

Our second contribution, inspired by DDP approaches \cite{ozaki2018stochastic}, is to show how our proposed method for state prediction using the Expansion-Compression Unscented Transform can be used with a gradient-descent approach based on Sequential Quadratic Programming (SQP) for \emph{online} policy optimization. This is unlike previous approaches that perform policy optimization offline using nonlinear optimization solvers. More specifically, our sigma point-based distribution propagation allows us to develop a model predictive framework that can be implemented as a computational graph. Backpropagating through this graph then allows us to use the policy gradients for policy optimization. Note that for being able to define the policy gradients, we use UT methods \cite{julier1997new, ebeigbe2021generalized} that admit analytical expressions and are differentiable almost everywhere except for the case when the covariance matrix is singular. Applying SQP on an optimization problem that models the computational graph obtained from sigma point propagation also allows us to pose our framework as an online controller tuning mechanism for autonomous robots. While our state prediction scheme can be used with any nonlinear optimizer to obtain an optimal policy similar to other approaches, we present the SQP approach as an alternative when an online policy update is desirable. This allows us to use model-based RL as a online tuning mechanism for controllers which is needed in the face of uncertainties and time-varying constraints.
%and use available solvers that  and not just as an offline policy search like previous methods[][].
%We also use policy itself is structurally defined by a control barrier function (CBF) based parametric quadratic program(QP)[].

This paper builds upon our earlier work in \cite{parwana2022recursive}, in which we introduced a preliminary version of the online parameter optimization framework for \emph{deterministic} systems. Compared to the conference version, the current paper includes the method of handling \emph{uncertain} dynamics, which corresponds to the introduction of the novel, data-efficient state prediction method that we call the Expansion-Compression Unscented Transform. The proposed expansion-contraction layers of our UT-based algorithm constitute, to the best of our knowledge, an important and standalone contribution to state prediction for online policy optimization under uncertainty, and the major contribution of the current paper. The UT-based method does not appear in \cite{parwana2022recursive}. The case studies also presented in this paper are different compared to the case studies in \cite{parwana2022recursive}.

\subsection{Paper Organization}

The rest of the paper is organized as follows. Section \ref{section::problem_formulation} describes the class of systems considered and the problem statement. Section \ref{section::methodology} introduces our methods for future-state prediction, as well as for policy optimization. 
%We also devise an \emph{online} parameter adaptation policy under state and input constraints and develop a constraint-aware gradient-descent update rule that aims to tune the parameters for prolonging the feasibility of the optimization problem while enhancing safety of the system trajectories. 
Finally, Section \ref{section::results} shows our implementation results. We compare the empirical performance of our expansion-compression prediction method against Monte Carlo simulations and a successive approximation method for \emph{state-dependent} distributions.
%and also show state-of-the-art results on the benchmark, the cart pole swing-up problem, at a much lower computational cost.  
Next, we show an application of our expansion-compression layer to solve a stochastic nonlinear MPC problem with IPOPT which is an off-the-shelf available solver.
 Finally, we show the efficacy of our framework for online tuning of policy parameters on two different case studies. In the first case study, we consider a leader-follower problem, where the follower's policy is computed as the solution to a Quadratic Program (QP) subject to multiple Control Barrier Function (CBF) constraints and input constraints.
 %; in fact, the proposed tuning method provides an extension to our earlier work in \cite{breeden2021highrelative}, in which CBF constraints and their coefficients/parameters are only manually defined and tuned. 
 In the second case study, we use our algorithm for a quadrotor stabilization task on safe-control-gym \cite{yuan2022safe} and compare its performance and computational characteristics with the baseline of GP-MPC \cite{hewing2019cautious}. 

\section{Problem Formulation}
\label{section::problem_formulation}

\subsection{Notations}
\label{section::notations}
The set of real numbers is denoted as $\reals$, non-negative numbers as $\reals^+$, and non-negative integers as $\mathbb{Z}^+$. For $x\in \reals$, $|x|$ denotes the absolute value of $x$.
%, $||y||$ denotes the $L_2$ norm of $y$. The interior and boundary of a set $\C$ are denoted by $\textup{Int}\mathcal{C}$ and $\partial \C$. 
%A continuous function $\alpha:[0,a)\rightarrow [0,\infty]$ for $a>0$ is called a $\classK$ function if it is strictly increasing and $\alpha(0)=0$. It is called $\classKinf$ if $a = \infty$ and $\lim_{a\rightarrow\infty} \alpha(r)=\infty$.
$\nabla_x$ denotes the partial derivative operator w.r.t variable $x$. The probability of an event $A$ happening is given by $P(A)$. $P(A|B)$ is the conditional probability of event $A$ given that event $B$ has happened. For $y\in \reals^n$, $y\sim D$ denotes that the random variable $y$ is distributed according to the distribution $D$. %\textbf{DO YOU REALLY USE THIS ANYWHERE?} The $k^{th}$ moment of $y$ is denoted as $M^k_D$. 
The first moment, namely expectation, and the second moment, namely covariance of what a random $x\in\mathbb R^n$ are denoted as $\Ep(x)$ or $\mu(x)$, and $\Sigma(x)$ respectively. A multivariate Gaussian with mean $\mu$ and covariance $\Sigma$ is given as $N(\mu,\Sigma)$. With $\Sigma_i$ we denote the $i$-th column of the covariance matrix $\Sigma$.

\subsection{Problem Statement and Approach}
\label{section::system_description}
Consider a nonlinear dynamical system with state $x \in \mathcal{X}\subset \reals^n $, control input $u\in \mathcal{U} \subset \reals^m$. The discrete-time state dynamics is given by
\eqn{
  x_{t+1} = f(x_t,u_t)
  \label{eq::dynamics_nonlinear_original}
}
where $t\in \mathbb{Z}^+$ denotes the time index and $f:\reals^n \times \reals^m \rightarrow \reals^n$ is a deterministic function for the system dynamics. Upon the application of a state-feedback policy $\pi:\mathbb R^n\times \mathbb R^\kappa \rightarrow \mathbb R^m$, so that $u_t = \pi(x_t, \theta)$, where $\theta\in\mathbb R^\kappa$ is a parameter vector, the closed-loop discrete-time dynamics are denoted as: 
\begin{align}
x_{t+1} = f_c(x_t),
\end{align}
where $f_c:\reals^n \rightarrow \reals^n$. When $f_c$ is unknown, we assume that the state transition model
\begin{align}
x_{t+1}\sim D_c(x_t)
\label{eq::dynamical_distribution}
\end{align}
is available, where $x_t$ is the current state, $x_{t+1}$ is the predicted state, $t\in \mathbb{Z}^+$ denotes the time index, and $D_c:\mathbb R^n \rightarrow \mathbb R^n$ is a nonlinear probability distribution,
%and $D:\reals^n \times \reals^m \rightarrow \reals^n$ is a distribution dependent on $x_t$ and $u_t$. We assume that $D$ is given in terms of its moments functions $M^i_D, i\in \{1,2,.., \infty \}$, each of which is a function of $x_t,u_t$. %For sake of consistency with the RL literature, we will also be denoting the state transition probability $\p(x_{t+1}|x_t,u_t)$ with $\mathcal{T}$. 
which has been obtained via model-learning algorithms such as Gaussian Process (GP) or Bayesian Neural Network. We assume that the model-learning algorithm is accurate enough, or in other words, that the true dynamics is a sample of the learned model, $f_c(x_t)\sim D_c(x_t)$. 

We furthermore consider $K$ constrained sets $\mathcal C_i \subset \mathcal{X}$ defined as the zero-superlevel sets of corresponding smooth constraint functions $c_i: \mathcal{X}\rightarrow \reals, i\in \{1,2,..,K\}$, such that 
\eqn{
        \C_i & \triangleq \{ x \in \X \;|\; c_i(x) \geq 0 \} \label{eq::superlevel-set1}.
}
For a random variable $x_t\in\mathcal X$ governed by the distribution in \eqref{eq::dynamical_distribution}, and for a control action $u_t\in\mathcal U$, we consider that each state-action pair $(x_t,u_t)$ results in a reward from the environment, given by the function $r: \X \times \U \rightarrow \reals$.
A policy $\pi:\X \rightarrow \U$ that maximizes the expectation of the cumulative reward for $t\in\{0,1,2,\dots,H\}$ while forcing the probabilistic satisfaction of the constraints at each time step $t$ is given as:
\begin{subequations}
\begin{align}
    \pi^* = \argmax_{u_t \in \mathcal{U}} \quad & \Ep \left(\sum_{t=0}^{H} r(x_t,u_t) \right), \label{eq::rl_objective} \\
    \textrm{s.t.} \quad %&\Ep[c_i(\hat x_t)] \geq 0 \label{eq::rl_constraint}, \; \\
    &P[c_i( x_t)\geq 0]\geq 1-\epsilon, \label{eq::rl_constraint}  \\
    & \forall i\in \{1,..,K\},  \forall t \in \{0,1,\dots,H\} \nonumber
    \end{align}
    \label{eq::rl}
\end{subequations}
where $u_t=\pi(x_t)$ and $\epsilon\in[0,1]$ is a user-defined risk. Note that the policy $u_t$ can be a parameterized policy in terms of parameters $\theta \in \reals^{\kappa}$, such as a neural network or a standard PD controller in which the P and D gains as viewed as tunable parameters. In that case, the optimization \eqref{eq::rl} is reformulated in terms of $\theta$ as shown below.

Since optimization over an infinite horizon (i.e., for $H=\infty$) is computationally infeasible, we seek to develop receding-horizon policies that satisfy the given constraints with high probability, as well as a method for state prediction that is computationally-efficient for implementation along with the online policy optimization. In summary, we consider the following problem: 

\textbf{Problem Statement:} Given a system modeled as in \eqref{eq::dynamics_nonlinear_original}, and a control policy $\pi:\mathcal X\times \mathbb R^\kappa \rightarrow \mathcal U$ such that $u_t = \pi(x_t,\theta)$, we seek to 

\noindent (\textbf{P.1}) find the parameter $\theta^*$ that maximizes the future expected reward over a finite time horizon $H$ subject to probabilistic safety constraints, i.e., solves for 
\begin{subequations}
\begin{align}
    \theta^* = \argmax_{\theta} \quad & \Ep \left(\sum_{\tau=0}^{H} r(x_{t+\tau}, u_{t+\tau}) \right), \label{eq::rl_objective} \\
     \textrm{s.t.} \quad & P\left[c_i(x_{t+\tau})\geq 0\right]\geq 1-\epsilon, \label{eq::rl_constraint}\\ 
 &  \forall i\in \{1,..,K\},  \forall \tau \in \{0,1,\dots,H\} \nonumber
    \end{align}
    \label{eq::rl_receding horizon}
\end{subequations}
\noindent where $u_{t+\tau}=\pi(x_{t+\tau};\theta)$, $\tau$ is the prediction time step, and $x_{t+\tau}$ is the predicted future state at time $t+\tau$ governed by the distribution $D_c$ \eqref{eq::dynamical_distribution}, and

\noindent (\textbf{P.2}) develop a method for obtaining computationally-efficient predictions of $x_{t+\tau}$ under the distribution $D_c$ \eqref{eq::dynamical_distribution}. 

\vspace{2pt}

\textbf{Approach:} To solve \textbf{P.2} we propose a novel prediction technique based on Unscented Transform (UT) that is accurate and data-efficient due to the proposed Expansion-Compression (EC) layer. To solve \textbf{P.1}, we use the proposed Expansion-Contraction layer to convert the stochastic optimization problem (5) to a deterministic optimization problem that can be solved by any available nonlinear solver like IPOPT. Additionally, when offline optimization is too expensive to perform, or when it does not lead to good solutions within a fixed number of solver iterations, we propose an online policy update algorithm that is inspired by Sequential Quadratic Programming and employs backpropagation through the computational graph generated by the states of the dynamical system \eqref{eq::dynamical_distribution}. The overall method, called FORESEE, comprising the prediction and online optimization phases, is described in detail in Section \ref{section::methodology}. In the next subsections we provide some preliminaries that help establish the notation and framework for FORESEE.

\subsection{State Distribution Prediction and the Unscented Transform}

Starting from a given state $x_t\sim D_c$ at time $t$, the state distribution at the next time step is obtained by marginalizing \eqref{eq::dynamical_distribution} over $p(x_t)$:
\eqn{
  p(x_{t+1}) = \int p( x_{t+1} | x_t; D_c )p(x_t) d x_t.
  \label{eq::state_prediction}
}
The above expression cannot be evaluated analytically for most distributions. When $p(x_t)$  and $p(x_{t+1} | x_t; D_c)$ are assumed to result from a Gaussian distribution whose mean and covariance are given by a Gaussian Process (GP) with a Gaussian kernel, PILCO \cite{deisenroth2011pilco} derived analytical formulas for computing the exact mean and covariance of $x_{t+1}$. Note that a choice of non-gaussian kernel in GPs would invalidate the formula derived in \cite{deisenroth2011pilco}. Another method is to perform Monte Carlo which, in theory, gives the exact distribution of $x_{t+1}$ in the limit of the number of particles going to infinity.

A sampling-based method to predict the distribution of $x_{t+1}$ in \eqref{eq::state_prediction} is via the Unscented Transform, which is used in Section \ref{sec::expansion-compression} to the Expansion-Conmpression version for data efficiency. At time $t$, for $x_t\in\mathbb R^n$, consider a set $\s_t$ of $N$ samples, or sigma points, $\s_{t,i}$, $i=1,\dots,N$, denoted $\s_t=\{\s_{t,1},\s_{t,2},\dots,\s_{t,N}\}$, and the set $w_t=\{w_{t,1},w_{t,2},\dots,w_{t, N}\}$ comprising their associated weights $w_{t,i}$. The sample (or empirical) mean and covariance of $x_t$ are computed as follows:
\begin{subequations}
\begin{align}
    \Ep_s[x_t] &= \sum_{i=1}^N w_{t,i}\s_{t,i}, \\ 
    \Sigma_s[x_t] &= \sum_{i=1}^{N} w_{t,i} (\s_{t,i}-\Ep_s[x_t])(\s_{t,i}-\Ep_s[x_t])^T.
\end{align}
\label{eq::sigma_point_basic}
\end{subequations}
Given the moments of a distribution of $x_t$, UT provides a principled way of selecting the sigma points $\mathcal{S}_t, w_t$ such that their empirical mean, covariance, and possibly other higher-order moments are equal to the moments of the given distribution. The standard UT algorithm \cite{julier1997new} is presented in Algorithm \ref{algo::UT} and, for a given mean and covariance of a $n$-dimensional distribution, it selects the best $N=2n+1$ sigma points $\s_{t,i}$ and weights $w_{t,i}$, $, i\in\{1,2,\dots,2n+1\}$, whose empirical mean and covariance given by \eqref{eq::sigma_point_basic} match the given moments. Note that the optimal choice of the free parameter $k$ in Algorithm \ref{algo::UT} depends on the original distribution. In practice, this is usually computed for specific distributions beforehand. For a priori unknown distributions, sigma point adaptation algorithms can be used, see \cite{turner2012model} for more information.
Other variations of UT such as Generalized Unscented Transform \cite{ebeigbe2021generalized} are capable of maintaining higher-order moments, and can also be used in our work.

% \st{These samples can also be used to approximate functions of random variables.  The integral of a function $f(x)$ can be approximated as}
% \eqn{
%     \int_a^b f(x) dx \approx \sum_{i=1}^N w_i f(\s_{[i]} \textrm{will remove this}
%     \label{eq::ut_evaluate_function}
% }
 %The greater representational power of UT is also one reason why it is considered a generalization over EKF for filtering applications \cite{wan2000unscented}.

\begin{algorithm}[ht]
    \caption{generate\_UT\_points}
    \begin{algorithmic}[1]
    \Require $\mu, \Sigma$ \Comment mean, covariance of the distribution
    \State Choose a free parameter $k>0$
    \State Calculate $2n+1$ sigma points 
        \begin{itemize}
            \item $\s_0 = \mu$
            \item $\s_{i} = \mu + \sqrt{(n+k)\Sigma_{i}}, \s_{i+n} = \mu - \sqrt{(n+k)\Sigma_{i}}$
        \end{itemize}
        for $i\in \{1,..,n\}$
    \State Calculate weights for the sigma points
        \begin{itemize}
            \item $w_0 = k/(n+k)$
            \item $w_i = 1/(2(n+k)), ~w_{i+n} = 1/(2(n+k))$
        \end{itemize}
        for $i\in \{1,..,n\}$
    \State Return $\s, w$
    \end{algorithmic}
    \label{algo::UT}
    \end{algorithm}

\subsection{Sequential Quadratic Programming}
\label{section::SQP}
Finally, we briefly describe Sequential Quadratic Programming (SQP) \cite{boggs1995sequential,tits2009feasible} that we use extensively in Section \ref{section::gd} for online, receding-horizon parameter adaptation. Consider an optimization problem in $z\in \reals^n$ with cost function $J:\reals^n\rightarrow \reals$, and constraint function $c:\reals^n \rightarrow \reals^m$ 
\begin{subequations}
    \begin{align}
    \min_z \quad & J(z), \\
    \textrm{s.t.} \quad & c(z)\leq 0.
    \end{align}
    \label{eq::sqp_intro}
\end{subequations}
In SQP, given an estimate $\bar z$ of the solution of the optimization problem \eqref{eq::sqp_intro}, gradients or sensitivities of the cost and constraint functions at $\bar z$ are used to compute infinitesimal update directions $d$ so that new estimate becomes $\bar{z}^+ = \bar z + d$.
The update direction $d$ is computed based on two factors:

(1) The improving-cost direction, i.e., the negated of the gradient of the cost function $-\nabla J|_{z=\bar z}$. 

(2) The feasible directions, i.e., the  directions for which constraints that are satisfied at the current estimate $\bar z$ are also satisfied at the new estimate $\bar z^+$. The set of feasible directions $\mathcal{F}(\bar z)$ at a point $\bar z$ up to first-order approximation is:
\eqn{
   \mathcal{F}(\bar z) = \{ d~ | ~ c(\bar z) + d^T \nabla c|_{z=\bar z} \leq 0 \}.
}
 The set $\mathcal{F}$ need not be computed explicitly. The update direction $d$ is computed by projecting the improving-cost direction into the set of feasible directions by solving the following linear program
\begin{subequations}
    \begin{align}
\min_d & \quad d^T \nabla J_{z=\bar z} \\
    \textrm{s.t} & \quad c(\bar z) + d^T \nabla c_{z=\bar z}  \leq 0.
\end{align}
\end{subequations}    

 Note that the notion of feasible directions is needed because moving only in the direction of improving cost does not guarantee that if $c(\bar z)\leq 0$ then $c(\bar z^+)\leq 0$.

%\textbf{THIS IS NOT THE PROBLEM THAT YOU ARE ACTUALLY CONSIDERING. SECTIONS II AND III SHOULD MERGE TO DESCRIBE THE PROBLEM STATEMENT (WHICH TO ME SHOULD BE A COMBINATION OF THE STOCHASTIC MODEL IN SECTION II WITH THE CONSTRAINTS HERE IN SECTION III) WITH A CLEAR OBJECTIVE (EG FIND SIGMA POINTS THAT PRESERVE STATISTICS OF THE DISTRIBUTION) THE OBJECTIVE MUST BE SOMETHING THAT THEN THROUGH THE ANALYSIS IS CLEAR THAT YOU DID ACCOMPLISH IT. YOU CAN HAVE THE PRELIMINARIES IN A SUBSECTION AFTER THE PROBLEM STATEMENT OR IN A SEPARATE SECTION. THEN YOU CAN GO TO THE SECTION IV WHERE YOU DESCRIBE THE APPROACH.}

\section{Proposed Approach: FORESEE}
\label{section::methodology}
This section explains how we perform future state prediction by propagating sigma points across multiple time steps, compute the policy gradient, and finally update the policy parameters using constrained gradient descent to respect state and input constraints.

Algorithm \ref{algo::FORESEE} presents our method for policy optimization. First, we explain our method for future state prediction presented in lines 5-10 of Algorithm \ref{algo::FORESEE}. Constrained gradient descent implemented in lines 16 and 17 is then detailed in the next subsection. 

\begin{algorithm}[h!]
\caption{FORESEE}
\begin{algorithmic}[1]
    \Require $x_0$, H, $\beta, \pi_\theta$ \Comment{initial state, time horizon, learning rate for policy update, parameterized policy}
    \State Initialize $\mu_0 = \mathbb E[x_0]$
    % \While {True}
    \For {N = 1 to num\_trials}
        % \State Move system with random or best available policy
        % \State Learn state transition dynamics $D_c$ from recorded data
        \State J = 0 \Comment{ Cost function }
        \State $\s_0, w_0$ = generate\_UT\_points( $\mu_0, 0$ )
        \For {t = 1 to T} \Comment{Predicting future}
            \State $\s_{next}, w_{next} \leftarrow $ expand\_sigma\_point($\s_{t-1}, w_{t-1}, D_c(\cdot;\pi_\theta)$)
            % \State $u = \pi_\theta(\s)$
            \State $J \leftarrow J + \Ep(r(\s_{next},\pi(\s_{next})))$
            \State $\s_t,w_t \leftarrow \textup{compress\_sigma\_points}(\s_{next}, w_{next})$
        \EndFor
        % \If{Offline Optimization}
        % \State Update $\theta = \argmin J(x_0,\theta)$ with any optimizer
        \If{Online Update}
        \State Find update direction $d_\theta$ using \eqref{eq::constrained_GD} or \eqref{eq::constrained_GD_feasibility}
        \State Update $\theta \leftarrow \theta + \beta d_\theta$
        \EndIf
    % \EndWhile
    \EndFor
\end{algorithmic}
\label{algo::FORESEE}
\end{algorithm}

\subsection{State Prediction with Expansion-Compression UT}
\label{sec::expansion-compression}
In this section we present our approach to solving Problem \textbf{P.2}. We first provide a high-level overview of the algorithm and then describe each individual operation in detail. 

Given $N$ sigma points and weights  $\s_t,w_t$ representing the state $x_t$ at time $t$, line 6 predicts the sigma points representing the state $x_{t+1}$ based on the dynamics function $D_c(x_t)$. Each sigma point $\s_{t,i}$ is mapped to a distribution  $D_c(\s_{t,i})$ based on dynamics $x_{t+1}|x_t = D_c(x_t)$ and Algorithm \ref{algo::UT} is used to sample new sigma points from this distribution. This leads to an increase in the total number of sigma points that is not scalable for multiple time-step prediction. Therefore, after evaluating the cost in Line 7, Line 8 performs an operation based on Algorithm \ref{algo::compress_sigma_points} to compress the sigma points $\s_{new}$ to obtain $N$ new sigma points. The procedure is repeated for $H$ time steps. \\
\begin{algorithm}[ht!]
    \caption{expand\_sigma\_points}
    \begin{algorithmic}[1]
    \Require $\s, w, D_c()$ \Comment{Collection of sigma points and their weights, closed-loop dynamics distribution}
    \State Initialize $\s_{new} = \{\}, w_{new} = \{\} $ 
    \For {i in size($\s$) }
        \State mean, cov = moments($D_c(\s_{i})$)
        \State $s', w'$ = generate\_UT\_points( mean, cov, skewness, kurtosis ) %\Comment{(Algorithm \ref{algo::UT})}
        \State $\s_{new}.\textup{append}(s')$
        \State $w_{new}.\textup{append}(w' * w_i)$ 
    \EndFor
    \State Return $\s_{new}, w_{new}$
    \end{algorithmic}
    \label{algo::expand_sigma_points}%
\end{algorithm}%

% \begin{assumption}
%     For sigma points and weights $\s_{t}, w_t$, the sample moments of $\hat x_t$ defined in \eqref{eq::sigma_point_basic} are equal to the true moments of distribution $\hat x_t$
% \end{assumption}

\noindent \textbf{Expansion Layer} Here we describe our sigma-point expansion layer in Algorithm \ref{algo::expand_sigma_points}. For every sigma point $\s_{t,i}$ of $x_t$, we obtain the moments of the distribution of $D(\s_{t,i})$. Using these moments, for each sigma point $\s_{t,i}$, we use the UT algorithm in line 4 to generate $N'$ new sigma points and weights denoted as $\s_{t,i}^j$ and $w^{j}_{t,i}$, $j\in \{1,.., N'\}$, $N'=2n+1$ for UT, as illustrated in Fig. \ref{fig::uncertainty_prop}. The weight of new sigma points $\s_{t,i}^j$ is obtained equal to $w_{t,i}^j\;w_{t,i}$, and the resulting $NN'$ new sigma points are returned. The next theorem shows that this scheme correctly represents the true distribution of $x_{t+1}$ up to mean and covariance.
% The extension to higher-order moments is skipped for brevity but can be proved in a similar fashion.

% \begin{theorem}
%  \label{theorem::expansion_layer}
%   Suppose the sigma points $\s_{t,i}, w_{t,i},i\in\{1,2.,..,N\}$ have sample moments equal to moments of random variable $x_t\in \mathbb R^n$. For each sigma point $\s_{t,i}$, consider the $N'$ new sigma points and weights denoted by $\s_{t,i}^j, w_{t,i}^j$, where $j\in\{1,2.,,.N'\}$ that have the same sample moments equal to distribution $D_c(\s_{t,i})$. Then the set of points $\s_{t,i}^j$ with corresponding weights $w_{t,i} w_{t,i}^j$ have a sample mean and covariance equal to those of the true distribution of $x_{t+1}$, $\Ep[x_{t+1}] = \Ep_s[x_{t+1}]$.
%   %with $E_s$ defined in \eqref{eq::sigma_point_basic} with points $\s_i^j, w_{t,i}w_{t,i}^j$.
%  \end{theorem}
% %Since $\s,w$ have the same moments as original distribution, we have $f(x)$
% \noindent 
% \begin{proof}
% See Appendix \ref{sec:proof-of-theorem}.
% \end{proof}

\begin{theorem}
 \label{theorem::expansion_contraction_layer}
  Suppose the sigma points $\s_{t,i}, w_{t,i},i\in\{1,2,..,N\}$ have sample moments equal to moments of random variable $x_t\in \mathbb R^n$. For each sigma point $\s_{t,i}$, consider the $N'$ new sigma points and weights denoted by $\s_{t,i}^j, w_{t,i}^j$, where $j\in\{1,2.,,.N'\}$ that have the same sample moments equal to distribution $D_c(\s_{t,i})$. Then, 
  (1) the set of points $\s_{t, i}^j$ with corresponding weights $w_{t, i} w_{t, i}^j$ have a sample mean and covariance equal to those of the true distribution of $x_{t+1}$, $\Ep[x_{t+1}] = \Ep_s[x_{t+1}]$. (2) Furthermore, suppose there exists $N$ sigma points $\s_{t+1,i}, w_{t+1,i}, i\in \{1,2,..,N\}$ whose sample moments are equal to moments of $x_{t+1}$. Then, the sigma points $\s_{t+1,i},w_{t+1,i}$ and $\s_{t, i}^j,w_{t, i}^j$ have the same sample mean and covariance.
  %with $E_s$ defined in \eqref{eq::sigma_point_basic} with points $\s_i^j, w_{t,i}w_{t,i}^j$.
 \end{theorem}
%Since $\s,w$ have the same moments as original distribution, we have $f(x)$
\noindent 
\begin{proof}
See Appendix \ref{sec:proof-of-theorem} for proof of the first argument. A visualization is also provided in Fig. \ref{fig::uncertainty_prop} where the weights of new sigma point $\s_{t, i}^j$ are obtained by multiplying the weight of the root node representing $\s_{t, i}$ with the weights of the edges that form $\s_{t, i}^j$. The second argument now trivially follows since both the set of sigma points, $\s_{t+1, i}, \;w_{t+1, i}$ (given) and $\s_{t, i}^j, \;w_{t, i}^j$ (from first argument), are known to have sample mean and covariance equal to that of $x_{t+1}$ and therefore equal to each other. The second argument forms the basis for the compression layer which is presented in the next section.
\end{proof}

Note that Theorem \ref{theorem::expansion_contraction_layer} does not assume a specific distribution, but relies on the assumption that $\s_i^j, w_i^j$ represent the true distribution. In this regard, our approach is not limited to GPs with Gaussian kernel as in PILCO \cite{deisenroth2011pilco} and other studies that use PILCO-inspired state prediction \cite{roberts2020safepilco, kamthe2018data, theodorou2010stochastic, pan2014probabilistic}. %but instead \textbf{HAVE YOU TESTED THIS? DO YOU HAVE ANY SOLID GROUND TO ARGUE THIS OR ARE YOUR JUST GUESSING?} can use GPs with any kernel, or also any other prediction algorithm such as Bayesian Neural Network.

\begin{algorithm}[h!]
    \caption{compress\_sigma\_points}
    \begin{algorithmic}[1]
    \Require $\s, w$ \Comment{Collection of sigma points and weights}
    \State $\mu_s, \Sigma_s$  = Sample mean, sample covariance (sigma points)
    \State Return generate\_UT\_points($\mu_s, \Sigma_s$)
    \end{algorithmic}
    \label{algo::compress_sigma_points}%
\end{algorithm}%

\noindent \textbf{Compression Layer}
Line 1 of Algorithm \ref{algo::compress_sigma_points} first computes the sample moments of $NN'$ sigma points $\s_{t,i}^j$ and weights $ w_{t,i}w_{t,i}^j$. Then using these moments, we generate new $N$ sigma points in line 2 using UT. 
%\textbf{OK CAN YOU SAY SOMETHING FORMAL HERE? SOMETHING ABOUT THE ACCURACY OF THE ESTIMATE AFTER THE EXPANSION AND THE COMPRESSION LAYER? YOU SHOULD FORM A THEOREM HERE AND MAYBE HAVE WHAT YOU CURRENTLY CALL A THEOREM TURN INTO A LEMMA.}

\begin{comment}
\textbf{I AM JUST REMOVING REMARK 2. IT IS CONFUSING AND DOESN'T REALLY MAKE ANY BETTER CLAIM/STORY FOR YOUR METHOD.}
\begin{remark}
    Defining Unscented Transforms with sigma points that preserve higher-order moments is an active area of research \cite{easley2021higher}. In the next section, we will also need the expansion and compression layers to be differentiable functions, i.e., the output of these layers should be differentiable w.r.t input to these layers. For UT methods that we employ in our work \cite{julier1997new, ebeigbe2021generalized}, the aforementioned operations have analytical expressions that are differentiable almost everywhere except for the case when the covariance matrix is singular. But the same does not hold for every available UT method. Nevertheless, any UT algorithm with differentiable operations can be used in our algorithm.
\end{remark}
\end{comment}

\begin{remark}
    Note that for the same number of samples, the expansion and compression operations are computationally more expensive than performing Monte Carlo simulations. The performance boost comes though from the fact that for the same accuracy level, only tens of UT sigma points are equivalent to having thousands or millions of MC particles. %\textbf{THE LATTER AGAIN SOUNDS LIKE A HYPOTHESIS SO I JUST MOVED IT TO FUTURE WORK. YOU HAVE MORE REMARKS THAN ACTUAL RESULTS IN THIS SECTION. THIS DOES NOT CREATE A POSITIVE IMPRESSION TO THE REVIEWER.} 
\end{remark}

\subsection{Constraint-Aware Gradient Descent}
\label{section::gd}
In this section, we propose a gradient-descent approach for (lines 16 and 17 of Algorithm \ref{algo::FORESEE}) for solving Problem \textbf{P.1}. Without loss generality, we divide the online policy update design into two cases and denote the prediction time with $\tau$.

\noindent \textbf{Case 1:} At time $t$, suppose that the policy $\pi(x_{t+\tau} ,\theta)$ given out of \eqref{eq::rl_receding horizon} satisfies the constraints \eqref{eq::rl_constraint} over the time steps $\tau=0,1,2,\dots,H$, but is suboptimal w.r.t. the performance objective. For $$R = \sum_{\tau=0}^H r( x_{t+\tau},\pi(x_{t+\tau},\theta)),$$ the standard RL update 
$$\theta^+ = \theta - \beta (\nabla_\theta \Ep[R]),$$ 
where $\theta^+\in \mathbb R^\kappa$ is the updated parameter at time $t$, $\beta\in[0,1]$ is the learning rate, and $\nabla_\theta \Ep[R] \in\mathbb R^{\kappa}$ is the gradient of the expectation of the reward, does not ensure that the policy $\pi( x_{t+\tau};\theta^+)$ continues to satisfy \eqref{eq::rl_constraint} for all $\tau\in\{0,1,2,\dots,H\}$. Therefore, inspired by Sequential Quadratic Programming \cite{tits2009feasible} and DDP literature \cite{andersson2015model, van2020online}, and based on our previous work in \cite{parwana2022recursive}, the following constrained gradient descent update rule is used:
\begin{subequations}%
        \begin{align}
        \theta^+ &= \theta + \beta d_\theta,\\
    d_\theta &= \argmax_d ~ d^T \nabla_\theta \Ep[R] =
    \argmin_d ~ -d^T \nabla_\theta \Ep[R] \\
    &\mbox{s.t.} \quad P[c_i( x_{t+\tau}) + d^T \nabla_\theta c_i( x_{t+\tau}) \geq 0]\geq 1-\epsilon,\label{eq::chance constraint_GD} \\
    & \forall \tau \in \{0,1,2,..,H\}, \forall i\in\{1,2,..,K\}\nonumber
        \end{align}
\label{eq::constrained_GD}%
\end{subequations}
where $\theta^+$ is the updated parameter at time $t$, $\beta\in[0,1]$ is the learning rate, $\epsilon\in(0,1]$ is the risk tolerance, $\nabla_\theta \Ep[R]$, is the gradient (wrt $\theta$) of the expectation of the reward, i.e., a deterministic value, but $\nabla_\theta c_i(x_{t+\tau})$ and $c_i(x_{t+\tau})$ are functions of the random variable $x_{t+\tau}$. Note that the independent variables are the parameter vector $\theta$ and the initial state $x_t$. Therefore, at each time $t+\tau$, the gradient $\nabla_\theta \Ep$ involves backpropagating through the graph. The above update rule forces the new parameter $\theta^+$ to satisfy the constraints that $\theta$ was satisfying upto first-order approximation about the current trajectory. 

We note that \eqref{eq::chance constraint_GD} is a probabilistic, or chance, constraint. Chance-constrained optimization problems are in general NP-hard, and only some cases are known to have analytical solutions.\footnote{E.g., the case where the cost and the constraints are linear in the decision variables, and their coefficients are Gaussian distributed vectors with known mean and covariance, and $\epsilon\leq 0.5$.} When the chance-constrained problem is intractable, several approximations (safe, convex, tractable approximations) of the original problem are usually considered, such as in \cite{nemirovski2007, nemirovski2009}. We articulate two such approaches to approximate constraints \eqref{eq::rl_constraint} and \eqref{eq::chance constraint_GD} using existing approaches in Appendix \ref{appendix::b} and \ref{appendix::c}.  Notably, the approximate constraint \eqref{eq::approximate-constraint-appendix} captures, consistently with one's intuition, the statistics of the state prediction, and the behavior of the constraint functions around the estimates of the state and the parameter vectors. In our ensuing though, for the sake of computational efficiency (reasons mentioned in Appendix \eqref{appendix::b}), we employ the approximation in \eqref{eq::gaussian_CF} from Appendix \ref{appendix::c}. 
%constraints Here we adopt such a formulation, which is articulated in Appendix \ref{sec:chance-constraint-approximation}. 

\begin{remark}
    While SQP provides aforementioned guarantees only upto first order approximation about current trajectory, Feasible SQP\cite{tits2009feasible} algorithms, although more computationally expensive, can also be used to provide guarantees up to higher orders. In practice, choosing a small $\beta$ helps in ensuring that the first-order approximation is valid.
\end{remark}
\noindent \textbf{Case 2:} At time $t$, suppose that the policy $\pi(x_{t+\tau} ,\theta)$ given out of \eqref{eq::rl_receding horizon} satisfies the constraints \eqref{eq::rl_constraint} over the time steps $\tau=0,1,\dots,H-1$, but not at the $\tau=H$ time step when predicting the future states $x_{t+\tau}, \tau\in\{0,1,..,H\}$. In this case, we relax the $j$-th constraint function at the $H$ time step so that $c_j(x_{t+H},u_{t+H})=-\delta_j, \; j\in \mathcal{J}$, where $\mathcal{J}$ is the set of constraints that could not be satisfied in the original constraint \eqref{eq::rl_constraint} and $\delta_J>0$ are called slack variables and are minimized through an objective function that, for example, minimizes $\delta_J^2$ instead of the original objective \eqref{eq::rl_objective}. $-\delta_J$ is also referred to as the infeasibility margin. Let $\delta_H = [\delta_{\mathcal{J}_1}, \delta_{\mathcal{J}_2}...,]$ be the vector of all slack variables. The update rule is then designed to be:
\begin{subequations}%
        \begin{align}
             \theta^+ &= \theta + \beta d_\theta, \quad\textrm{where}\\
    d_\theta  &= \argmax_d ~ -d^T \nabla_\theta \Ep[{\delta_H}^T \delta_H] \\
    ~ \textrm{s.t.} &~ P[c_i(x_{t+\tau}) + d^T \nabla_\theta c_i(x_{t+\tau})\geq 0] \geq 1-\epsilon,
   % & \quad \quad \textrm{s.t.}  ~\Ep_{x_t}(e(x_t,u_t,\theta)) + \nabla_\theta \Ep_{x_t}(e(x_t,u_t,\theta)) d \geq 0 \nonumber \\
    \\
    & \quad \quad \quad ~ \forall \tau\in \{0,1,..,H-1\}, \forall i\in \{1,2,..,K\} \nonumber \\
    & ~P[c_j(x_{t+H}) + d^T \nabla_\theta c_j(x_{t+H})\geq 0] \geq 1-\epsilon\\
    & \quad \quad \quad . \nonumber \forall j\in \{1,2,..,K\}\setminus \mathcal{J} 
        \end{align}
\label{eq::constrained_GD_feasibility}
\end{subequations}
The above update rule minimizes the slack, and as a result the infeasibility margin, while enforcing that the constraints that were already satisfied continue to be satisfied up to first-order approximation. The approximation of the probabilistic constraints with robust ones can be done as in the earlier case.
\begin{remark}
    Since the optimizations  \eqref{eq::constrained_GD},\eqref{eq::constrained_GD_feasibility} are linear programs, they can be solved efficiently. Therefore in practice, we can perform at least one step of gradient descent at each time $t$ or multiple steps if solving to improve feasibility as in \eqref{eq::constrained_GD_feasibility}. Finally, a desirable property for online policy optimization is convergence in fixed time. While convergence guarantees of SQP have been examined in literature\cite{nocedal1999numerical}\cite{liu2005global}\cite{dohrmann1997efficient}, we leave this analysis for future work.
\end{remark}
% \red{NOT HERE: Our leader-follower example in Section \ref{section::results} shows how to implement this.}
% The above update guarantees that the new parameter $\theta^+$ will lead to feasible solutions of the QP policy. 

\vspace{-1mm}
\section{Results}
\label{section::results}

We present a number of case studies to show the efficacy of the proposed method. The first case study demonstrates the accuracy and sample efficiency of the Expansion-Compression UT compared to state-of-the art methods on uncertainty propagation; the second case study demonstrates the application of FORESEE in offline trajectory optimization, while the third and fourth show the applicability of the method for online controller tuning. The code and videos for our simulations can be found at \href{https://github.com/hardikparwana/FORESEE}{https://github.com/hardikparwana/FORESEE}.

\subsection{Benchmark: Accuracy of State Prediction}
We compare the accuracy of our state distribution prediction scheme to Monte Carlo and PILCO. Consider the following dynamics:
\eqn{
x_{t+1} & \sim D(x_t), \\
D(x) &= N( g(x), 0.01+\textup{diag}(g(x)^2) ), \\
  g(x) &= a \begin{bmatrix}
      \cos(x(1))+0.01 \\
      \sin(x(2))+0.01
  \end{bmatrix}, 
%  \\
% x_{t+1} \sim f(x_t) \\
% a = 10\\ 
% a = 20
}
where $x=\begin{bmatrix}x(1) & x(2)\end{bmatrix}^T$ and $a\in\mathbb R$. Note that this function is such that even Monte Carlo performs poorly in accurately predicting the  distribution over multiple time steps when the number of samples is small. Starting at time $t=1$ with state $x_1 = \begin{bmatrix}0 &0\end{bmatrix}^T$, we forward propagate the state two times to predict the value of the state $x_3$ at time $t=3$. To evaluate the performance, we perform Monte Carlo with 500, 5000, and 50,000 particles as well as PILCO-like successive Gaussian approximation type of update for mean and covariance matrix. We also perform two versions of our proposed algorithm: one with successive expansion operations only, leading to a surge in the number of sigma points, and another with successive expansion-compression operations that keep the number of sigma points constant. The performance of the compression-expansion layer is evaluated when using two different UT algorithms:  vanilla UT (Algorithm \ref{algo::UT}) and generalized UT\cite{ebeigbe2021generalized}. The free parameter $k$ in Algorithm \ref{algo::UT} is chosen to be 1 which is known to be optimal for Gaussian distributions.

\begin{figure}
\begin{subfigure}[b]{0.24\textwidth}
     \centering \includegraphics[width=0.99\linewidth]{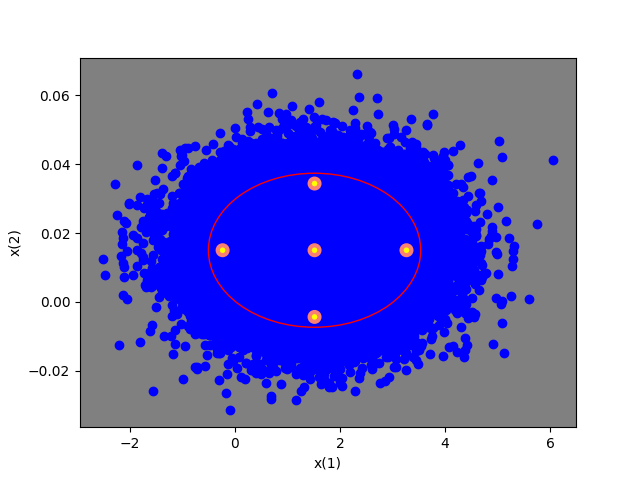} 
     \caption{\small{Horizon = 1, UT with EC}} 
    \label{fig::fig1}%
\end{subfigure}   
\begin{subfigure}[b]{0.24\textwidth}
     \centering \includegraphics[width=0.99\linewidth]{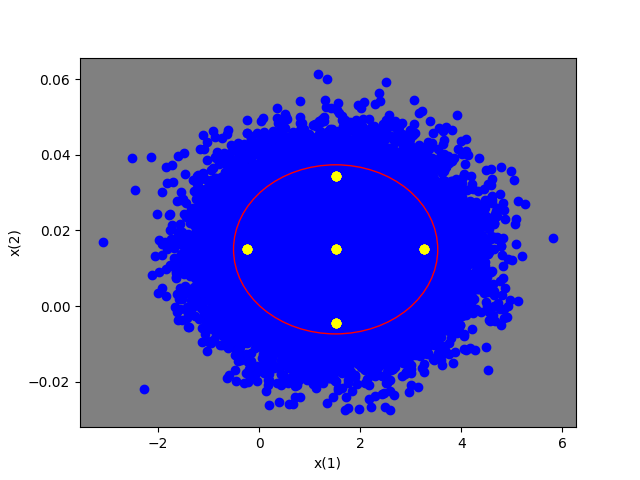} 
     \caption{\small{Horizon = 1, UT with E only}} 
    \label{fig::fig2}%
\end{subfigure}   
\begin{subfigure}[b]{0.24\textwidth}
     \centering \includegraphics[width=0.99\linewidth]{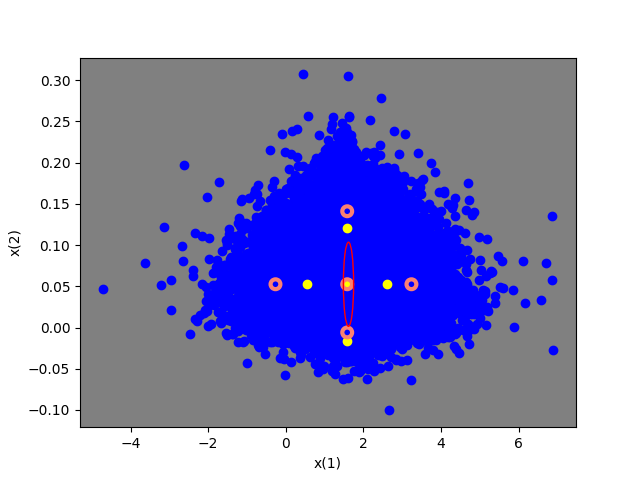} 
     \caption{\small{Horizon = 2, UT with EC}} 
    \label{fig::fig7}%
\end{subfigure}   
\begin{subfigure}[b]{0.24\textwidth}
     \centering \includegraphics[width=0.99\linewidth]{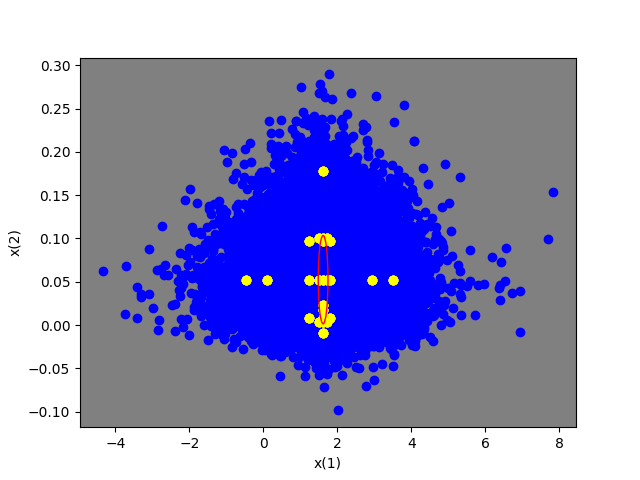} 
     \caption{\small{Horizon = 2, UT with E only}} 
    \label{fig::fig8}%
\end{subfigure} 
\begin{subfigure}[b]{0.24\textwidth}
     \centering \includegraphics[width=0.99\linewidth]{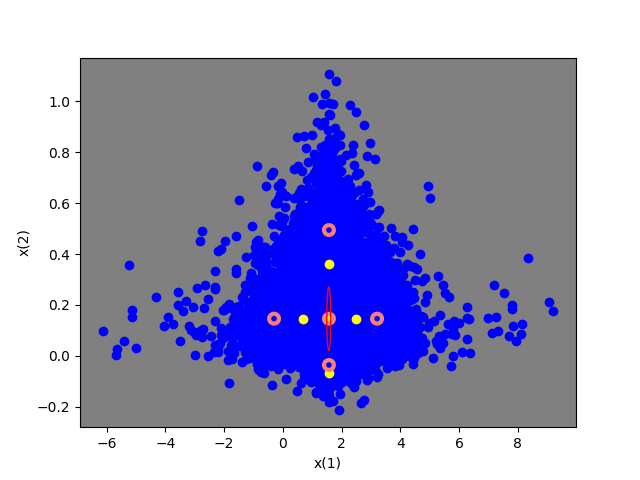} 
     \caption{\small{Horizon = 3, UT with EC}} 
    \label{fig::fig13}%
\end{subfigure}   
\begin{subfigure}[b]{0.24\textwidth}
     \centering \includegraphics[width=0.99\linewidth]{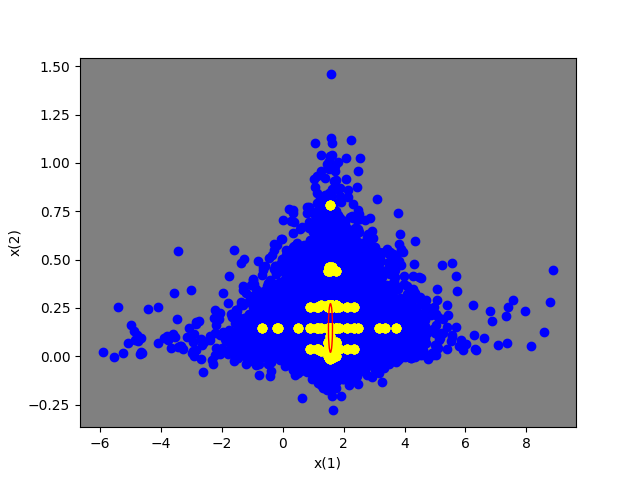} 
     \caption{\small{Horizon = 3, UT with E only}} 
    \label{fig::fig14}%
\end{subfigure}   
\begin{subfigure}[b]{0.24\textwidth}
     \centering \includegraphics[width=0.99\linewidth]{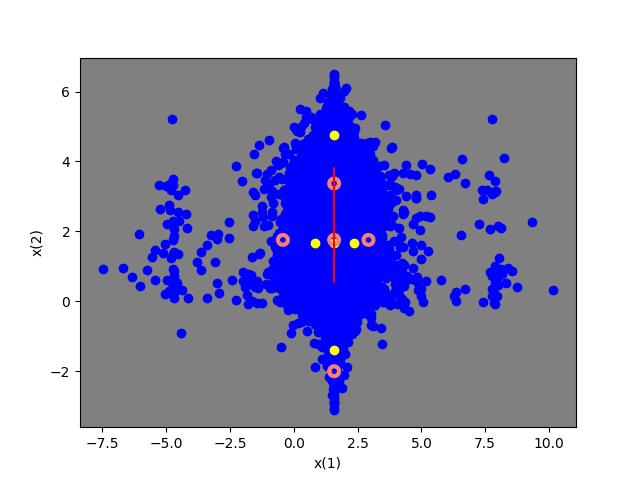} 
     \caption{\small{Horizon = 6, UT with EC}} 
    \label{fig::fig24}%
\end{subfigure}   
\begin{subfigure}[b]{0.24\textwidth}
     \centering \includegraphics[width=0.99\linewidth]{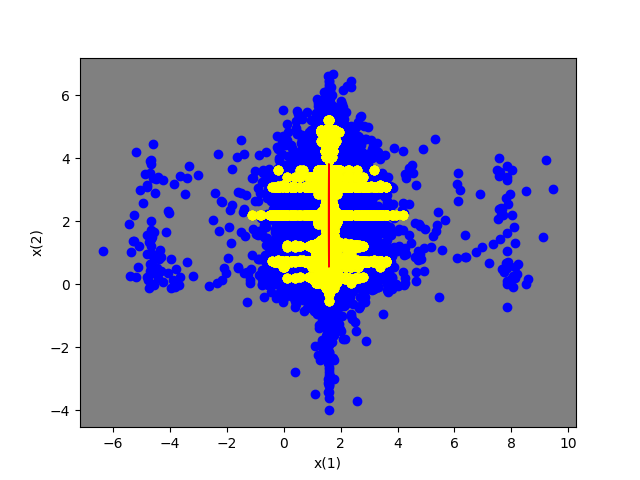} 
     \caption{\small{Horizon = 6, UT with E only}} 
    \label{fig::fig23}%
\end{subfigure}   
\begin{subfigure}[b]{0.24\textwidth}
     \centering \includegraphics[width=0.99\linewidth]{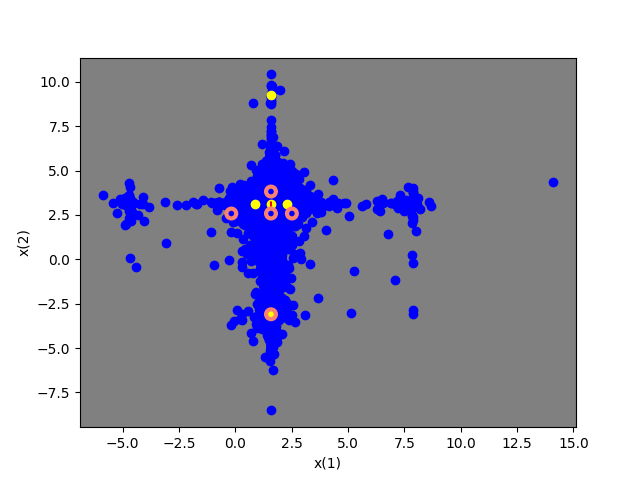} 
     \caption{\small{Horizon = 8, UT with EC}} 
    \label{fig::fig23}%
\end{subfigure}   
\begin{subfigure}[b]{0.24\textwidth}
     \centering \includegraphics[width=0.99\linewidth]{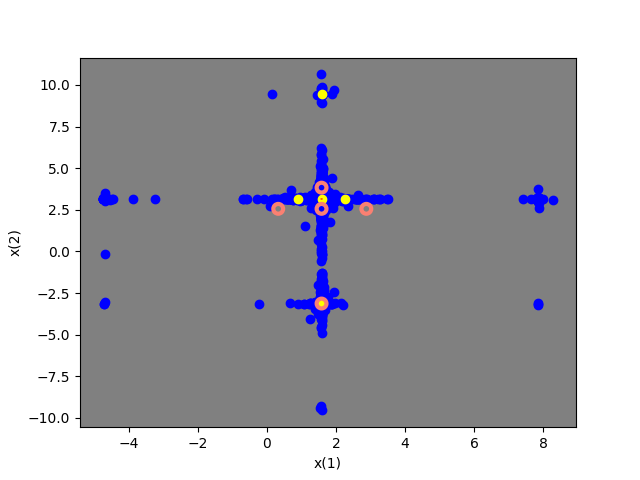} 
     \caption{\small{Horizon = 10, UT with EC}} 
    \label{fig::fig23}%
\end{subfigure}   
\caption{\small{The predicted state distribution under different methods: Blue samples correspond to the 50,000 MC particles. Yellow and salmon-colored samples are the sigma points generated using successive Expansion-Compression (EC) operations. Yellow samples employ UT (\ref{algo::UT}) and salmon samples employ Generalized UT \cite{ebeigbe2021generalized} in its expansion and compression operation. Red ellipse represents the 95\% confidence ellipse of the distribution obtained by successive Gaussian approximation.}} 
\label{fig::gaussian_nonlinear_dynamics}
\end{figure}

\begin{figure}
    \centering    
    \includegraphics[width=0.9\linewidth]{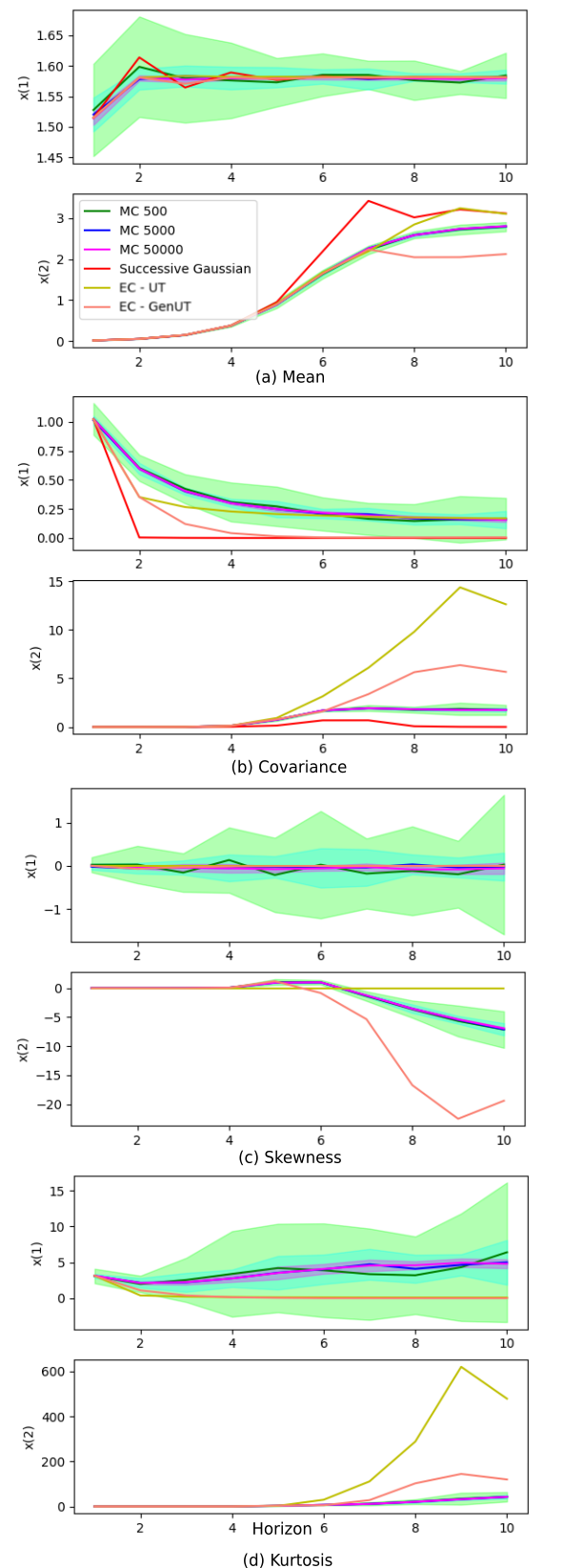}
    \caption{\small{First four moments with different methods: Monte Carlo (MC), succcessive Gaussian, Expansion-Compression (EC) with UT, and with Generalized UT (GenUT). The successive Gaussian approximation is only shown for the first two moments as it is identically zero and constant for skewness and kurtosis. The shaded areas show two standard deviation variations in MC runs.} }%
    \label{fig::uncertainty_compare}%
\end{figure}

\begin{figure}
    \centering    
    \includegraphics[width=1.0\linewidth]{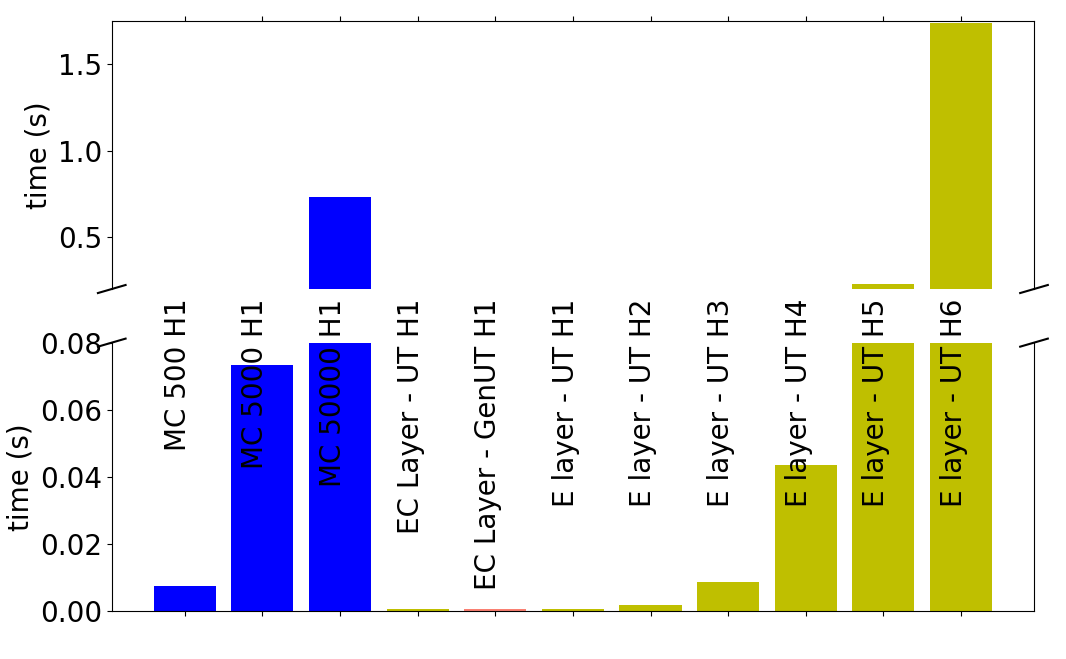}
    \caption{\small{Time taken by each of methods in a naive Python implementation (without any optimizations): Monte Carlo (MC), Expansion-Compression (EC), and Expansion-only (E). All operations are performed on a single core of CPU. H value denotes the horizon. The time per step for MC and EC is fixed. The time taken per step by E Layer increases with horizon as the number of particles increases.}}%
    \label{fig::uncertainty_compare_time}%
\end{figure}

Fig. \ref{fig::gaussian_nonlinear_dynamics} shows the predicted state distributions obtained under different methods (proposed, Monte Carlo, and successive Gaussian approximation) for $a=20$. Fig. \ref{fig::uncertainty_compare} shows the corresponding predicted moments for different values for $a=20$ and Fig. \ref{fig::uncertainty_compare_time} shows the computation times for a naive Python implementation. %\textbf{THAT IS NOT REALLY SHOWN IN THE DATA: EITHER YOU GIVE SPECIFIC NUMBERS OR REMOVE} Note that MC's accuracy improves drastically as the number of particles is increased and is therefore very sensitive to the number of particles unless a large number of particles are used. 
Monte Carlo (MC) simulations are run 20 times and sample sizes of 500 and 5000 are observed to have high variance in their estimates.  Successive Gaussian approximation performs poorly in all scenarios and its covariance identically tends to zero. It also exhibits oscillatory behavior in Fig. \ref{fig::uncertainty_compare}(a). On the other hand, in agreement with previous works \cite{ebeigbe2021generalized}, we note that UT based methods lead to a similar accuracy as that of MC with large number of particles, but at a much lower computational cost. Also, as evident from Fig. \ref{fig::gaussian_nonlinear_dynamics} and \ref{fig::uncertainty_compare_time}, UT with expansion only quickly becomes non-scalable, as within a horizon of 6, the number of sigma points is much more than the MC particles, and hence is no longer computationally cheap. On the other hand, the proposed UT with the Expansion-Compression (EC) layer keeps the number of sigma points fixed across a time step, and is computationally the most efficient method. Finally, generalized UT \cite{ebeigbe2021generalized} leads to more accurate estimates of moments than vanilla UT in general. Vanilla UT is also unable to convey information on the asymmetry of the resulting distributions. Generalized UT on the other hand chooses sigma points asymmetrically to preserve the diagonal elements of skewness and kurtosis moment matrices of the sigma points in compression operation and is therefore better represents the true distribution.

\begin{figure}
\begin{subfigure}[b]{0.24\textwidth}
     \centering \includegraphics[width=1.0\linewidth]{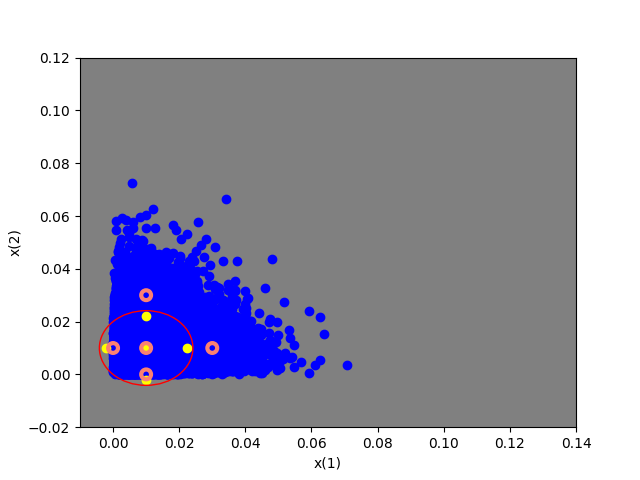} 
     \caption{\small{Horizon = 1, UT with Expansion-Contraction}} 
    \label{fig::fig1}%
\end{subfigure}   
\begin{subfigure}[b]{0.24\textwidth}
     \centering \includegraphics[width=1.0\linewidth]{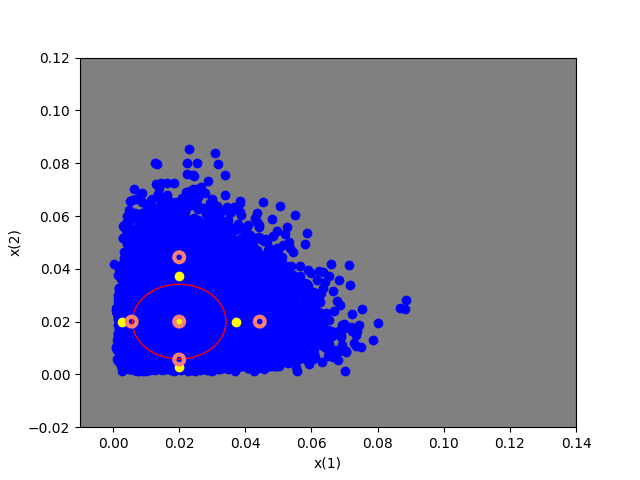} 
     \caption{\small{Horizon = 2, UT with Expansion-Contraction}} 
    \label{fig::fig2}%
\end{subfigure}   
\begin{subfigure}[b]{0.24\textwidth}
     \centering \includegraphics[width=1.0\linewidth]{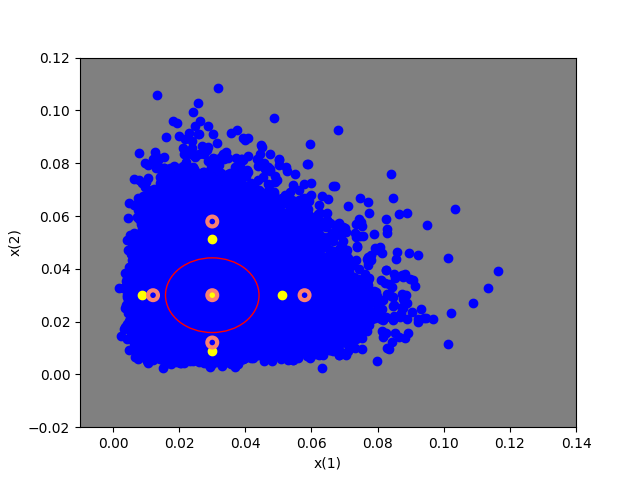} 
     \caption{\small{Horizon = 3, UT with Expansion-Contraction}} 
    \label{fig::fig7}%
\end{subfigure}   
\begin{subfigure}[b]{0.24\textwidth}
     \centering \includegraphics[width=1.0\linewidth]{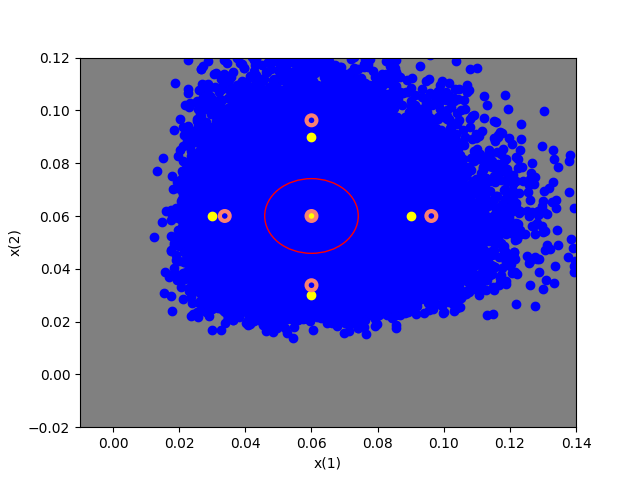} 
     \caption{\small{Horizon = 6, UT with Expansion-Contraction}} 
    \label{fig::fig8}%
\end{subfigure} 
\caption{\small{The predicted state distribution under different methods for dynamics given by gamma distribution in \eqref{eq::gamma_dynamics}. The colors of samples represent follow the notation in Fig. \ref{fig::gaussian_nonlinear_dynamics}.}} 
\label{fig::gamma_prediction}
\end{figure}

To further show the efficacy of our expansion-compression layers, we propagate states through a non-gaussian dynamics function - the gamma distribution. Given the shape and scale parameters, $k$ and $\beta$ respectively, the first four moments of gamma distribution $\Gamma$ are given by
\eqn{
\mu_\Sigma = k\beta, \Sigma = k\beta^2, S_\Sigma = \frac{2}{\sqrt{k}}, K_\Sigma = \frac{6}{k}+3
\label{eq::gamma_dynamics}
}
The predicted state distributions are shown in Fig. \ref{fig::gamma_prediction}. GenUT is again observed to better represent the asymmetrical characteristics of the true distribution.

\subsection{Application to Stochastic Trajectory Optimization}
\begin{figure}
\begin{subfigure}[b]{0.5\textwidth}
     \centering \includegraphics[width=0.93\linewidth]{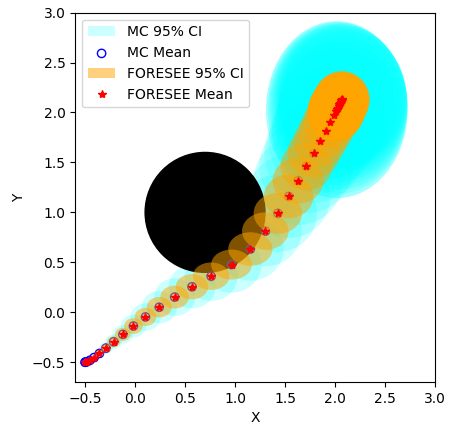} \caption{\small{Case 1: Constraint satisfaction in expectation}} 
    \label{fig::mpc_case1}%
\end{subfigure}   
\begin{subfigure}[b]{0.5\textwidth}
     \centering \includegraphics[width=0.93\linewidth]{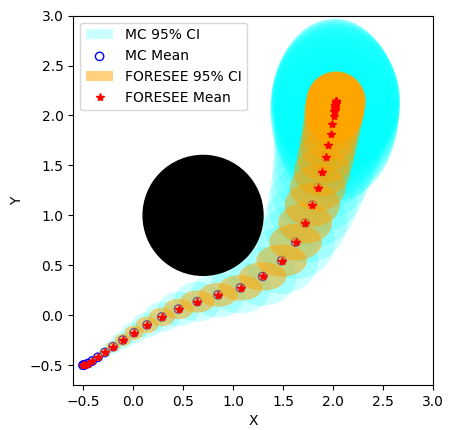} \caption{\small{Case 2: Constraint satisfaction in 95\% confidence interval}} 
    \label{fig::mpc_case2}%
\end{subfigure}   
\begin{subfigure}[b]{0.5\textwidth}
     \centering \includegraphics[width=0.93\linewidth]{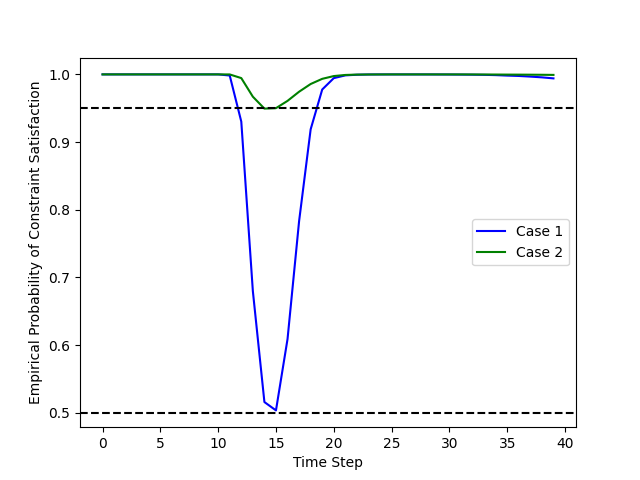} \caption{\small{Constraint satisfaction rate over time for Case 1 and 2.}} 
    \label{fig::mpc_constraint_violation.png}%
\end{subfigure}  
\caption{\small{The predicted trajectories of dynamic unicycle modeled robot. The start location is at (-0.5,-0.5) and the goal is at (2,2). The orange and blue ovals represent the 95\% confidence interval (w.r.t mean and covariance) of the sigma point and Monte Carlo states respectively.}} 
\label{fig::mpc_trajs}
\end{figure}

In this section, we show an application of state distribution prediction to stochastic trajectory optimization. Consider a dynamic unicycle model with state $x_t=[p_{xt} ~ p_{yt} ~ \psi_t ~ v_t]^T$ and the following uncertain dynamics
\eqn{
f(x_t) &= x_t + \begin{bmatrix}
    v_t\cos\psi_t \\ v_t\sin\psi_t \\ \omega_t \\ a_t
\end{bmatrix}\Delta t, \label{eq::du_mpc_dynamics}\\
x_{t+1} & \sim N\bigg( \frac{3}{2} f(x_t), \frac{\textrm{diag}[(f(x_t)-x_t)^2]}{10} \bigg)
}
where $p_{xt}, p_{yt}$ are the position coordinates along the axes of an inertial frame, $\psi_t$ is the heading angle, $v_t$ is the speed, $u_t=[\omega_t ~ a_t]^T$ is the control input consisting of angular velocity $\omega_t$ and acceleration $a_t$, and $N(\mu, \Sigma)$ is the normal distribution with mean $\mu$ and covariance matrix $\Sigma$. The objective is to navigate the robot to its goal location while avoiding collision with a circular obstacle as shown in Fig. \ref{fig::mpc_trajs}. Denote the distance to the obstacle at time $t$ as $d_t$, and the obstacle radius as $d_{min}$, and consider two cases: the first enforces collision avoidance, a scalar constraint, in expectation, that is
\eqn{
\Ep[{d_t}^2 - d_{min}^2] \geq 0 \label{eq::mpc_case1}
}
and the second enforces collision avoidance with 95\% confidence interval (CI) defined as 
\eqn{
 \Ep[{d_t}^2 - d_{min}^2] - 2 \sqrt{\Sigma[{d_t}^2 - d^2_{min}] } \geq 0.
 \label{eq::mpc_case2}
}
% \eqn{
%     CI(d) = \{ 
% z \in \reals ~| ~\Ep[d]-2 \sqrt{\Sigma[d]} \leq z \leq \Ep[d] + 2 \sqrt{\Sigma{d}} \}
% }
The objective function in \eqref{eq::rl} is chosen as
\eqn{
r(x_t, u_t) = 10 \left(\bigl\| \begin{bmatrix}
    p_{xt} \\ p_{yt}
\end{bmatrix} - G \bigr \|\right)^2 + \|u_t\|^2
}
where $G$ is the goal location. We solve \eqref{eq::rl} for time step $\Delta t=0.05s$ in \eqref{eq::du_mpc_dynamics} and horizon $H=40$ using IPOPT solver. Fig. \ref{fig::mpc_case1} and \ref{fig::mpc_case2} show the resulting trajectory for both cases. To validate our method, we also perform Monte Carlo (MC) with 50,000 particles (to simulate the stochastic system dynamics), and propagate the state forward with the control input designed using expansion-compression layers (our method). Fig \ref{fig::mpc_constraint_violation.png} shows MC's empirical probability of constraint satisfaction with time. We observe that our method provides good tracking of the mean of the state while under-approximating its covariance, see Fig. \ref{fig::mpc_case1}, \ref{fig::mpc_case2}. It is also noted that only a small violation of the probabilistic constraint is observed around time step 15, see Fig. \ref{fig::mpc_constraint_violation.png}, where the dotted black lines show the desired constraint satisfaction probability at 0.5 and 0.95 for Cases 1 and 2 respectively.
%Note that since the constraints \eqref{eq::mpc_case1}, \eqref{eq::mpc_case2} are nonlinear, the confidence interval of state distribution do not represent the confidence interval of distances to obstacle

\subsection{Application to Online Controller Tuning}

Controllers are prone to performance degradation in environments for which they were not optimized beforehand, especially when they are expected to satisfy several constraints. To further showcase the capability of FORESEE to perform online tuning of complex policies, we use the proposed framework as a tuning mechanism for control policies that are resulting as the solution to a Quadratic Programs (QP) subject to Control Barrier Function (CBF) conditions for safety \cite{parwana2022recursive, ames2016control}, termed thereafter the CBF-QP controller, since its parameters are difficult to generalize to new environments.

A continuously differentiable function $h:\mathcal{X} \rightarrow \reals$ is a CBF for the set $\mathcal{D} = \{x\in \mathcal{X} \;|\; h(x)\geq 0\}$ under the deterministic dynamics $x_{t+1} = F(x_t,u_t), ~F:\mathcal{X}\times \mathcal{U}\rightarrow \mathcal{X}$, $t\in\mathbb Z^+$, if there exists an extended class-$\mathcal{K}_\infty$ function $\nu$ such that
 \eqn{
 \sup_{u_t\in \mathcal{U}} \left[ h(x_{t+1}) - h(x_t) \right] \geq -\nu(h(x_t)), ~ \forall x_t \in \mathcal{X}, \forall t\in\mathbb Z^+.\label{eq::cbf_derivative}} 
For simplicity, we use $\nu(x)=\alpha x, \alpha\in \reals^+$, where $\alpha$ is a parameter that can be tuned. 
Given $K$ barrier functions $h_i: \mathbb R^n\rightarrow \mathbb R$ for $i\in \{1,2,..,K\}$, we can define corresponding CBF derivative conditions \eqref{eq::cbf_derivative} for each of the barriers $h_i$, each of which will include a parameter $\alpha_i$. Let also the desired state be encoded in terms of a Control Lyapunov Function (CLF) $V:\mathbb{R}^n\rightarrow \mathbb R$, and consider the policy $\pi_{QP}(x_t;{\alpha_0,\alpha_1,...\alpha_K})$ obtained by the CBF-CLF-QP:
\begin{subequations}
        \begin{align}
            \min_{u_t,\delta} \quad & J(u_t) = (u_t-u^d_{t})^TP(u_t-u^d_{t}) + Q\delta^2\\
            \textrm{s.t.} \quad &  V(x_{t+1}) \leq (1-\alpha_0)V(x_t) + \delta \\
                    & h_{i}(x_{t+1}) \geq (1-\alpha_{i})h_{i}(x_{t}) ~,\forall i \in \{1,..,K\} \nonumber\\
                    & \quad \quad \quad A u_t \leq b,  
        \end{align}
        \label{eq::CBF_MDP}
\end{subequations}
where $P\in \reals^{m\times m}, Q\in \reals^+$ are the weight factors, $u^d_t$ is the nominal (or desired) control input at time $t$, $\delta \in \reals$ is a slack variable used to relax the CLF condition, and $A,b$ specify the control inputs in the set $\mathcal{U}$. We will refer to $\theta = \begin{bmatrix}\alpha_0 & \dots & \alpha_K\end{bmatrix}^T$ as the parameter vector %(comprising the parameters $\alpha_0,\dots,\alpha_K$ involved in the CBF and CLF conditions) 
that needs to be learned. %Each CBF and CLF constraint in \eqref{eq::CBF_MDP} forms a constraint $c_i$ in \eqref{eq::rl}. 
Note that a manual tuning of the parameter $\theta$ to fixed values can make the constraints incompatible at future time steps (even if they are originally compatible), and therefore the QP \eqref{eq::CBF_MDP} will not have a solution. We, therefore, aim to achieve adaptation of the parameters over a time horizon to optimize the reward at the end of the time horizon, while the system states are propagated over this time horizon with our computationally-efficient UT method.

To illustrate the efficacy of the proposed state prediction and online policy optimization method, we consider two mobile robots moving in a leader-follower fashion, where the leader is modeled as a single integrator with dynamics as follows:
\eqn{
    \dot p_{x,l} = u_1, \quad \dot p_{y,l} = u_2,
}
where $p_{x,l}, p_{y,l}$ are the position coordinates of the leader with respect to an inertial reference frame, and $u_1, u_2$ are the velocity components along the axes of reference frame, 
whereas the follower is modeled with unicycle dynamics:
\eqn{
  \dot p_{x,f} &= u\cos \phi, ~ \dot p_{y,f} = u\sin \phi, ~ \dot \phi = \omega,
}
where $p_{x,f}, p_{y,f}$ are the position coordinates of the follower with respect to the same inertial reference frame, $\phi$ is the angle between the x-axis of the body-fixed reference frame of the follower and the inertial frame, and $u, \omega$ are the linear and angular velocity of the follower expressed in its body-fixed reference frame.

The follower has a limited camera Field-of-View (FoV), and its objective is to maintain the leader inside it, and preferably at the center of FoV. The follower's controller is given as a QP subject to three constraints %encoding the FoV: one each for minimum distance, maximum distance, and maximum angle w.r.t camera axis. More specifically, the three barrier conditions are defined 
as follows: 
\begin{subequations}
\begin{align}
h_1 &= s^2 - s_{min}^2\geq 0,\label{eq::barrier_colission}\\
h_2 &= s_{max}^2 - s^2 \geq 0,\\ 
h_3 &= b - \cos(\gamma) \geq 0,
\end{align}
\end{subequations}
where $s=\sqrt{(p_{x,l}-p_{x,f})^2+(p_{y,l}-p_{y,f})^2}$ is the Euclidean distance between the follower and the leader, $s_{min}$ is the minimum allowed distance for collision avoidance, $s_{max}$ is the maximum allowed distance for accurate detection using the onboard camera, $b$ is the bearing vector from the follower to the leader, and $\gamma$ is the FoV angle of the follower's camera. The CBF-CLF-QP \eqref{eq::CBF_MDP} uses the CBF conditions of the aforementioned constraints, the Lyapunov function 
\begin{align}
    V(s) = \left(s - \frac{s_{min}+s_{max}}{2} \right)^2,
\end{align}
and the input constraints $|u|\leq 2.5, |\omega|\leq 4$. 
 
The follower is assumed to have access to its own dynamics. The leader is moved independently with the following stochastic dynamics 
\begin{align}
\dot p_l &= N(\begin{bmatrix}1.25u_x\\1.25u_y\end{bmatrix},\begin{bmatrix}0.25({u_x}^2+u_y^2)&0\\0&0.25(u_x^2+{u_y}^2)\end{bmatrix}) \nonumber \\
u_x &= 2, u_y=3\cos 4\pi t
\label{leader_noisy_model}
\end{align}
% Regarding its knowledge of the leader's model, we consider that a noisy version of the velocity trajectories of the leader are available to the follower, given as: 
% where $u_1, u_2$ the nominal velocity components of the leader as defined earlier. 
The purpose of the case study is %twofold: First, to showcase the efficiency of the online state prediction and the online policy adaptation in the case when only a noisy model for the leader dynamics is known to the follower. Second, 
to showcase that the proposed method prolongs feasibility in the case of multiple constraints and in particular control input bounds, which in several cases render the problem \eqref{eq::CBF_MDP} infeasible. %Hence the ability for online tuning of the policy parameters offers increased performance from the perspective of not resorting to worst-case disturbances only, as in e.g., \cite{panagou2012icra}. 

%In our simulation implementations we chose to evaluate the nominal constraints with the mean values of the follower's states, i.e., in effect consider a very high risk $(\epsilon=1)$, since this preserves the form of the quadratic program \eqref{eq::CBF_MDP}. As demonstrated by the simulations, the uncertainty influence is in practice negligible, in part because the policy is implemented in a receding-horizon, state-feedback fashion. Alternatively, one can  consider the probabilistic formulation of CBF and CLF constraints in \cite{dhimantac2023, longlcss22}, or our robust control formulation for CBF and CLF constraints in \cite{gargacc21}.

% To evaluate the approach, we consider two cases, termed ``adaptive" and ``nominal adaptive", where the former corresponds to the follower considering the motion model \eqref{leader_noisy_model}, and the latter corresponds to the follower considering the mean value of the model above, but with zero covariance. Both cases employ FORESEE in a receding horizon fashion to predict state distributions and perform constrained gradient descent. In addition, for each case ``adaptive" and ``nominal adaptive" we consider two cases, namely solving \eqref{eq::CBF_MDP} without control input bounds, and solving \eqref{eq::CBF_MDP} with control input bounds. Videos of the simulation results are available at \textrm{https://github.com/hardikparwana/FORESEE}.

\begin{figure}
    \centering   
    \begin{subfigure}{1.00\linewidth}
    \includegraphics[width=0.95\columnwidth]{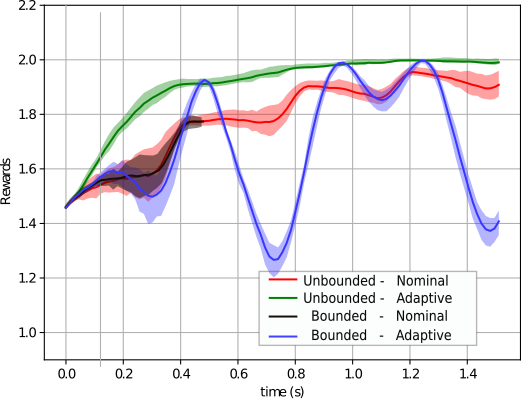}
    \caption{\small{Variation of reward with time.} }%
    \label{fig::lf_rewards_wob}%
\end{subfigure}
\begin{subfigure}{1.00\linewidth}
\includegraphics[width=0.95\columnwidth]{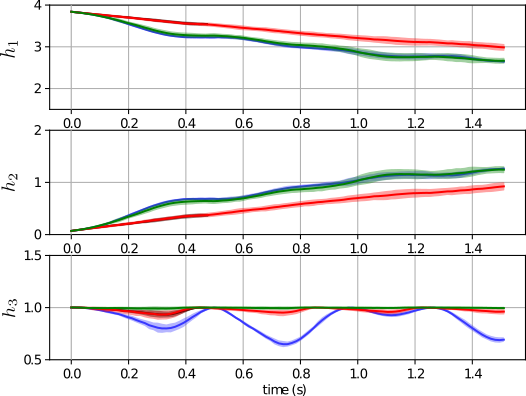}
    \caption{\small{Evolution of $h_1$ (minimum distance), $h_2$ (maximum distance) and $h_3$ (maximum angle w.r.t camera axis) with time $t$. Constraint maintenance corresponds to $h_i(t)>0, i\in\{1,2,3\}$, $\forall t\geq 0$.}} 
    \label{fig::lf_barriers_wob}%
\end{subfigure}
\begin{subfigure}{1.00\linewidth}
\includegraphics[width=0.95\linewidth]{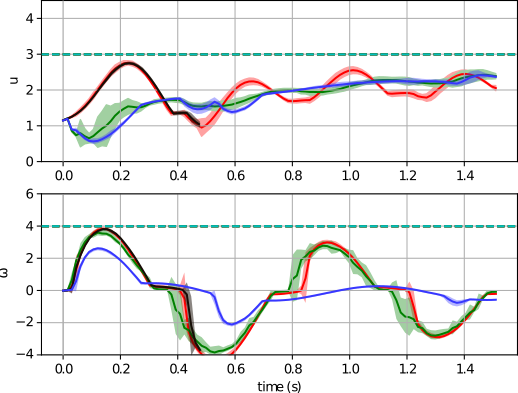}
    \caption{\small{Evolution of follower's control inputs with time.}} 
    \label{fig::lf_inputs_wob}%
\end{subfigure}
\caption{Responses over time for the system reward, barrier functions and the control input for the cases of bounded and unbounded control input and with nominal (fixed) and adaptive controller parameters. The adaptation is performed using Algorithm \ref{algo::FORESEE}}\label{fig::lf_results}
\end{figure}

Since the leader's state is 2-dimensional, we use 5 sigma points in the Unscented Transform in Algorithm \ref{algo::UT} in FORESEE to represent the leader's state. Fig. \ref{fig::lf_results} shows the results for unbounded and bounded control inputs, each with fixed parameter values (no adaptation) and with adaptive parameter values wherein we use the proposed online parameter tuning framework (Algorithm 1). Since the leader's true dynamics is stochastic, the results are reported over 30 trials.
%We observe that the adaptive case, in general, leads to lower variance in the follower's trajectories compared to the nominal adaptive case, which is expected as in the former case the follower is aware of the uncertainty in the leader's motion. The worst-case rewards and barrier functions are also larger valued for the adaptive case, implying better performance and safety, compared to the nominal adaptive case. 
Videos of the simulation results are available at \textrm{https://github.com/hardikparwana/FORESEE}

% In the unbounded control input case (Fig. \ref{fig::without_control_bounds}), we note that keeping the initial choice of parameters (fixed parameters) is able to satisfy constraints at all times (Fig. \ref{fig::lf_barriers_wob}), but with lower reward compared to "adaptive" and "nominal adaptive" (Fig. \ref{fig::lf_rewards_wob}); this means that the follower is not able to keep the leader centered in the FoV. 
In the unbounded control input case, we note that keeping the initial choice of parameters (fixed parameters) is able to satisfy constraints at all times (Fig. \ref{fig::lf_barriers_wob}), but with lower reward compared to "adaptive" case (Fig. \ref{fig::lf_rewards_wob}); this means that the follower is not able to keep the leader centered in the FoV. 

When control input bounds are enforced, it is noteworthy that in the fixed parameter case, the QP becomes infeasible after some time and the simulation fails, note the shorter black trajectories in Fig. \ref{fig::lf_results}. The update rule proposed in \eqref{eq::constrained_GD_feasibility} succeeds in maintaining feasibility and adapts the parameters so that QP is feasible and the constraints are satisfied (Fig. \ref{fig::lf_barriers_wob}). Interestingly, in the presence of input bounds, the learned policy realizes that the follower does not have enough control bandwidth to focus on performance, i.e., keep the leader at the center of the FoV and only ensures constraint satisfaction by turning minimally to keep the leader inside the FoV (i.e., angular velocity $\omega$ remains close to zero); this can be also verified from the corresponding control inputs in Fig. \ref{fig::lf_inputs_wob}.

\subsection{Benchmark Comparison}
\begin{figure}
\begin{subfigure}[b]{0.5\textwidth}
     \centering \includegraphics[width=0.92\linewidth]{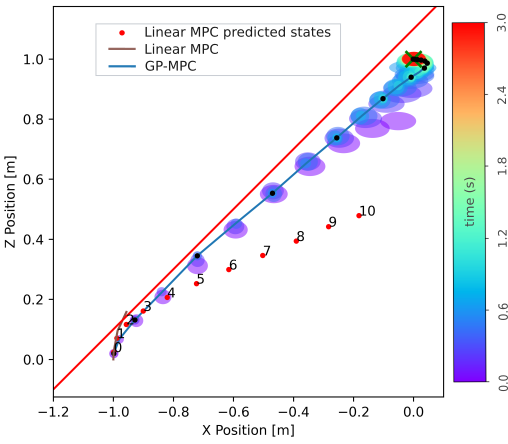} \caption{\small{GP-MPC and Linear MPC results from safe-control-gym\cite{yuan2022safe}. }} 
    \label{fig::gp_mpc}%
\end{subfigure}   
\begin{subfigure}[b]{0.5\textwidth}
     \centering
     \includegraphics[width=0.8\linewidth]{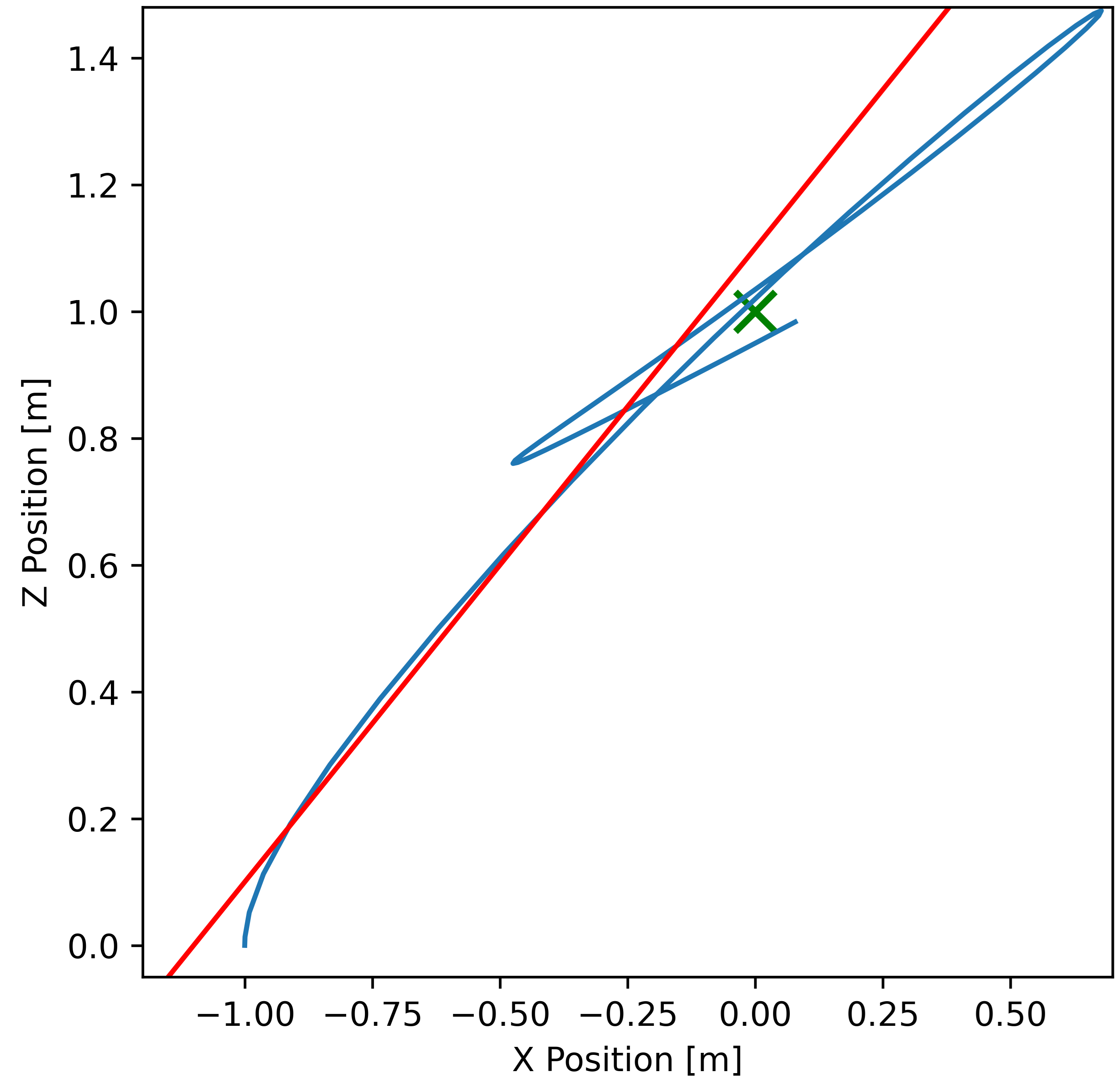} \caption{\small{Quadrotor trajectory with the nominal geometric controller.}} 
    \label{fig::my_nominal}%
\end{subfigure}
\begin{subfigure}[b]{0.5\textwidth}
     \centering
     \includegraphics[width=0.8\linewidth]{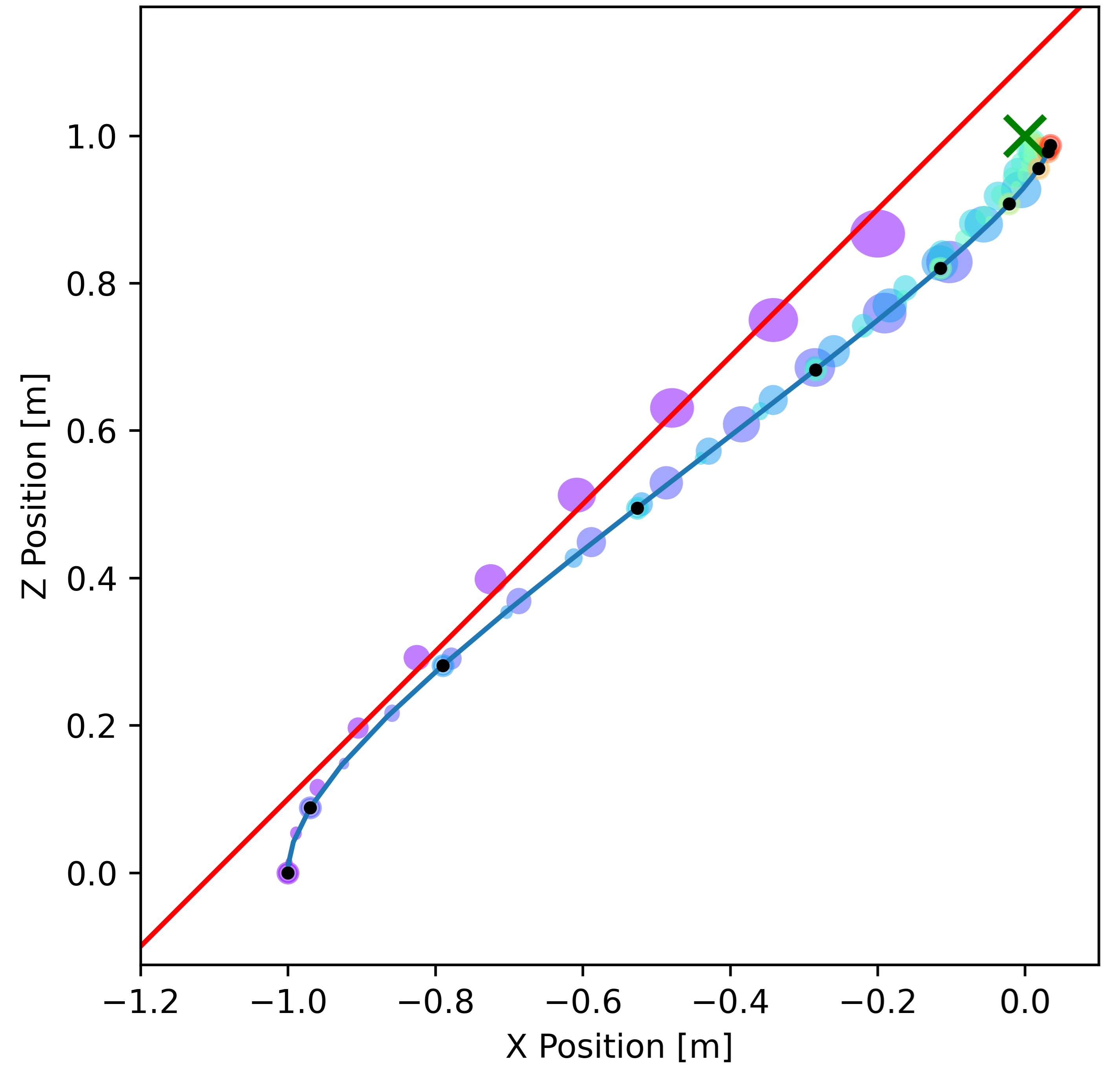} \caption{\small{Quadrotor trajectory with FORESEE-tuned geometric controller.}} 
    \label{fig::my_adaptive}%
\end{subfigure}
\caption{\small{Quadrotor stabilization results for benchmarking in safe-control-gym. The ellipses in (a) and (c) show the 95\% confidence interval Gaussian representing the states predicted by GP-MPC. The black dots are the (chosen) states from which predictions were made. The predicted ellipses originating from these dots have color decided according to the colormap corresponding to the timestamp of the black dot state.}} 
\label{fig::safe-control-gym}
\end{figure}

We compare our Algorithm \ref{algo::FORESEE} for online controller tuning  to the GP-MPC approach in \cite{hewing2019cautious} when applied on a 2D quadrotor model subjected to state and input constraints. We implement our framework in safe-control-gym \cite{yuan2022safe} that also provides implementation of \cite{hewing2019cautious}. The quadrotor dynamics is:
\begin{equation*}
\begin{matrix}
 \ddot p_x = \frac{(T_1+T_2)\cos\theta}{m} &
 \ddot p_z = \frac{(T_1+T_2)\sin\theta}{m} - g & \ddot \theta = \frac{(T_2-T_1)d}{I_y},
\end{matrix}
\end{equation*}
where $p=\begin{bmatrix}p_x & p_z\end{bmatrix}^T$ are the horizontal and vertical positions of the quadrotor, $\theta$ is the pitch angle, $T_1,T_2$ are the thrusts of left and right motors, and the control input is denoted as $u=\begin{bmatrix}T_1 & T_2\end{bmatrix}^T$. The quadrotor is subject to the state constraint $Ap \leq b$ with $A=[1 ~ -1], b=2$. We use the following nominal position controller:
\eqn{
a_d &= -k_x (p - p_d) - k_v (\dot p - \dot p_d) - k_{rx} \frac{A}{\|A\|} \tanh\left(\frac{1}{b-Ap}\right)
\label{eq::quad_controller}
}
where $T= \begin{bmatrix}
    \cos \theta  &  \sin\theta 
\end{bmatrix} (a_d + mg)$, $\theta_d = \tan^{-1} (T_x, T_z)$, $M= I_y ( - k_R (\theta - \theta_d) - k_{Rv} \dot\theta )$, $T_2 = 0.5 ( T + M/L )$, $T_1 = 0.5 ( T - M/L )$, and $a_d,p_d,\dot p_d$ are the desired accelerations, positions and velocities. $k_x, k_v, k_{rx}$ are gains to be tuned for stabilization performance and to satisfy the safety constraint at all times, and $k_{R}, k_{Rv}$ are constant gains to stabilize the quadrotor attitude control. Note that since the safe-control-gym benchmark runs the controller code at 10Hz, the resulting discretization errors are expected to degrade the performance of the continuous-time controller in \eqref{eq::quad_controller}. 
%However, as we will show, online adaptation of controller gains will help alleviate this degradation.
%along with additional small disturbances induced in safe-control-gym the continuous time controller in \eqref{eq::quad_controller} is implemented at 10 Hz, leading to discretization errors in addition to the disturbances induced in safe-control-gym implementation.
We perform all simulations on a Ubuntu 22.04 laptop running on an i9-13905H processor and no GPU acceleration is employed. We use the same trained Gaussian Process that GP-MPC uses to predict future states in our predictive framework. The prediction horizon for all the algorithms is 10. In Fig. \ref{fig::gp_mpc}, we show the results for GP-MPC and linear MPC controllers from safe-control-gym for a quadrotor stabilization task. The robot starts at $x=\begin{bmatrix}-1 & 0 & 0 & 0 & 0 & 0\end{bmatrix}^T$ and has to stabilize at $x=\begin{bmatrix}0 & 0 & 1 & 0 & 0 & 0\end{bmatrix}^T$. The simulation is run for 30 steps with a controller time step of 0.1 sec. The control input (thrust forces) are bounded, $0\leq T_i\leq 0.2 N, i\in\{1,2\}$. 
%Note that MPC is implemented in the receding horizon. 
The means and covariances of the system states for times $[t, t+10]$ predicted by MPC at time $t$ are shown and visualized by $95\%$ confidence interval ellipses. The colormap denotes the simulation time, ranging from violet ($t=0$) to red ($t=3$). 

\begin{figure}[h]
     \centering \includegraphics[width=0.95\linewidth]{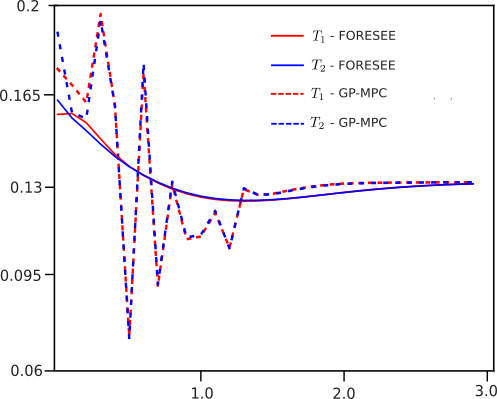} \caption{\small{Control input variation for GP-MPC and FORESEE.}} 
    \label{fig::control_comparison}%
\end{figure}

In contrast to MPC, we deploy FORESEE with online constrained gradient descent to tune the gains of nominal controller \eqref{eq::quad_controller}. The gains are initiated as $k_x=0.07, k_v=0.02, k_{rx}=0, k_{R}=60, k_{Rv}=10$ and mass and inertia are taken to be the true values from safe-control-gym. Fig. \ref{fig::my_nominal} shows the performance of the nominal controller for fixed values of the gains. Under this choice of gains, the system violates safety constraints and exhibits poor stabilization performance. 
%Note that the geometric controller is a continuous time controller so implementing it at 10Hz on a quadrotor with nonlinear dynamics is bound to break the convergence guarantees that it enjoys. 
However, with our online prediction-based tuning, we can tune parameters for constraint satisfaction and better stabilization as seen in Fig. \ref{fig::my_adaptive}. Fig. \ref{fig::control_comparison} also shows that our control inputs are smoother than that of GP-MPC. This can be attributed to the fact that our method incrementally changes the gains of a state-feedback controller, thereby avoiding large changes in control input. 

% Finally, while this example is supposed to last 3 minutes the receding horizon GP-MPC implementation takes 140 seconds whereas our algorithm (also implemented in receding horizon) takes only . 
safe-control-gym uses IPOPT solver for solving the nonlinear MPC optimization. It also employs some approximations proposed in \cite{hewing2019cautious} to impose linear chance constraints in the MPC to improve optimization solve time. The first run takes 19.8 seconds for IPOPT to find a solution. The subsequent runs benefit from setting the initial guess based on the solution of the last solved MPC (a standard trick in receding horizon implementations) and take between 2.7 to 8.5 seconds to find a solution. Note that on average, despite using the efficient IPOPT solver written in C++, the time to solve a single MPC (even with a good initial guess) is still more than the real-world duration of 3 seconds that this quadrotor maneuver is supposed to take.  FORESEE is written in PYTHON using the JAX library and JAX's autograd features are used to compute the required gradients. Each update in \eqref{eq::constrained_GD} and \eqref{eq::constrained_GD_feasibility} takes 0.02 seconds. The very first run takes 118 iterations of \eqref{eq::constrained_GD_feasibility} to find a feasible solution but after that, we limit the number of iterations in receding horizon implementation to 10 to finish the simulation within 3 seconds. Writing a more efficient C++ code to further speed up our computations will be performed in future work.

\section{Conclusion}
In this paper, we introduced a method, called FORESEE, for the online prediction and adaptation of uncertain nonlinear systems. The proposed method utilizes 1) the Unscented Transform and moment matching via an Expansion-Compression operation to propagate nonlinear uncertain distributions with comparable accuracy and much less computational demands than Monte Carlo simulations; 2) ideas from Sequential Quadratic Programming in order to perform online policy parameter optimization with constrained gradient-descent approaches. The case studies demonstrate the efficacy of the FORESEE in terms of computational efficiency, prediction accuracy, and online performance tuning compared to state-of-the-art approaches for state propagation, trajectory optimization and controller tuning. Future work will focus on deriving theoretical analysis on the accuracy of the expansion-compression operation, as well as on experimental implementations of the FORESEE algorithm on robotic setups.  %We considered in particular the problem of tuning online the parameters of Control Barrier Functions, and the simulations verified that our algorithm is able to find policies that respect state and input constraints. Unlike most of the previous state-of-the-art, we implemented our algorithm online in a receding horizon fashion. Future work will include how our method can be used on subgroups of MC particles to help reduce the number of samples in MC with higher representational power of UT. We also aim to incorporate simultaneous model learning and comparison with other state-of-the-art algorithms on benchmark problems to compare data sample efficiency, an evaluation of our method on real systems, and an analysis of the convergence rate of the algorithm and on the continuity of solutions.

% \newpage
% \bibliographystyle{IEEEtrans}
% \bibliography{icra.bib}

% \section*{Acknowledgments}

%% Use plainnat to work nicely with natbib. 
%\newpage
\bibliographystyle{IEEEtranS}
\bibliography{icra.bib}

\vspace{-2pt}
\appendix

\subsection{Proof of Theorem \ref{theorem::expansion_contraction_layer}}
\label{sec:proof-of-theorem}

%Before presenting the theorem's proof, we remind the reader of two properties related to the moments of a distribution. 
At time $t$, consider the random variable $\hat x_t$, which has to be propagated through a random function $D_c(\cdot)$ to obtain the random variable $\hat x_{t+1}$. The mean of $\hat x_{t+1}$ is:
\eqn{
  \Ep[\hat x_{t+1}] = \Ep[ D_c(\hat x_t) ] = \Ep \bigl[ \Ep[D_c({\hat x_{t}}) | \hat x_t ] \bigr], %= \Ep \bigl[ \mu(D_c(\hat x_t)) \bigr]
  \label{eq::mean_prop}
}
where we use the law of iterated expectation \cite{bertsekas2008introduction} in the second equality, while by the law of total variance \cite{bertsekas2008introduction} we have the second-order moment as:
\eqn{
  \Sigma[\hat x_{t+1}] &=  \Ep\bigl[  \Sigma [ \hat x_{t+1} | \hat x_t] \bigr]  + \Sigma \bigl[ \Ep[ \hat x_{t+1}|\hat x_t ] \bigr] \nonumber\\
  & = \Ep [ \Sigma(D_c(\hat x_t)| \hat x_t) ] + \Sigma[ \mu(D_c(\hat x_t)|\hat x_t) ]. 
  \label{eq::cov_prop}
  %\Ep  \bigl[ y y^T \bigr]          = \Ep \bigl[ \Ep[ yy^T | x_k ] \bigr] 
         %= \Ep \bigl[ \Sigma_{D_c}(x_k) \bigr]
}

\begin{proof}
    \noindent For brevity, we remove the explicit dependence on $t$ for sigma points and weights. For example, $\s_i$ stands for $\s_{t,i}$. The sample expectation defined by sigma points $\s_i^j$ and weights $w_iw_i^j$ is given as %that represent the state $\hat x_{t+1}$,
    \eqn{
        % \Ep_{s}[\hat x_{t+1}] &= \sum_i^{N} \sum_j^{N'} w_i w_i^j \s_i^j \\
        \Ep_{s}[\hat x_{t+1}] &= \sum_i^{N} \sum_j^{N'} w_i w_i^j \s_i^j = \sum_i^{N} w_{i} \sum_j^{N'} w_i^j \s_i^j% \nonumber
    }
    Since $\s_i^j,w_i^j$ are sigma points for $D_c(\s_i)$ and since we have assumed that they have sample moments equal to true moments of $D_c(\s_i)$, we have
%\stackrel{\mathclap{\normalfont\mbox{\footnotesize{\eqref{eq::sigma_point_basic}}}}}{=}
\eqn{
    \Ep[D_c(\s_{i})]& = \sum_j^{N'} w_i^j \s_i^j \label{eq::mu_sample} \\ %\mu_{D_c}(\s_{[i]}), 
    % \Sigma_{D_c}(\s_{[i]})
    \Sigma[D_c(\s_{i})]  & = \sum_j^{N'} w_i^j (\s_i^j - \Ep[D_c(\s_{i})] )(\s_i^j - \Ep[D_c(\s_{i})] )^T  \label{eq::sigma_sample}
}
Then, we have
    % \eqn{
    %      \Ep_{s}[\hat x_{t+1}] &= \sum_i^{N} \sum_j^{N'} w_i w_i^j \s_i^j = \sum_i^{N} w_{i} \sum_j^{N'} w_i^j \s_i^j \nonumber 
    %      }
         \eqn{
         \Ep_{s}[\hat x_{t+1}] &= \sum_i^{N} w_{i} \sum_j^{N'} w_i^j \s_i^j \\
&\stackrel{\mathclap{\normalfont\mbox{\footnotesize{\eqref{eq::mu_sample}}}}}{=} \sum_i^{N} w_{i} \; \Ep[D_c(\s_{i})] = \sum_i^N w_i \Ep[D_c(\hat x_t) | \hat x_t = \s_{i}]  \nonumber \\
& = \Ep[\Ep[D_c(\hat x_t) | \hat x_t = \s_{i}]] \stackrel{\mathclap{\normalfont\mbox{\footnotesize{\eqref{eq::mean_prop}}}}}{=} ~ \Ep[ \hat x_{t+1} ].
    }
Similarly, for sample covariance of sigma points $\s_i^j$ and weights $w_iw_i^j$
\eqn{
        & \Sigma_{s}[\hat x_{t+1}] = \sum_i\sum_j w_i w_i^j (\s_i^j-\Ep[\hat x_{t+1}])(\s_i^j-\Ep[\hat x_{t+1}])^T \nonumber \\
        & = \sum_i w_{i}\left( \sum_j w_i^j(\s_i^j-\Ep[\hat x_{t+1}])(\s_i^j-\Ep[\hat x_{t+1}])^T \right) %\nonumber 
        \\
%     }
% \eqn{
% }
% \eqn{
            & = \sum_i^{N} w_{i} \sum_j^{N'} w_i^j\left(  (\s_i^j - \Ep[D_c(\s_{i})]) (\s_i^j - \Ep[D_c(\s_{i})])^T \right) \nonumber \\
            %& \qquad \qquad \qquad  \textrm{(where } \mu_{D_c}(\s_{[i]})=\sum_j w_{j} \s_{[i,j]} \textrm{)} \nonumber \\
            & + \sum_i^{N} w_{i} \sum_j^{N'} w_i^j ( \Ep(D_c(\s_{i})) - \Ep[\hat x_{t+1}] )( \Ep(D_c(\s_{i})) - \Ep[\hat x_{t+1}] )^T \nonumber \\
            & + 2 \sum_i^{N} w_{i} \sum_j^{N'} w_i^j \left( \s_i^j - \Ep[D_c(\s_{i})] \right) \left( \Ep[D_c(\s_{i})] - \Ep[\hat x_{t+1}] \right)^T \label{eq::cov_computation} \\
            & \stackrel{\mathclap{\normalfont\mbox{\footnotesize{\eqref{eq::sigma_sample}}}}}{=} \sum_i^{N} w_{i} \Sigma[D_c(\s_{i})] \nonumber \\
            & +  \sum_i^{N} w_{i} \left( \Ep[D_c(\s_{i})] - \Ep[\hat x_{t+1}] \right)\left( \Ep[D_c(\s_{i})] - \Ep[\hat x_{t+1}] \right)^T  \nonumber + 0 \nonumber \\
            & = \Ep[\Sigma[D_c(\hat x_t)\big|\hat x_t = \s_{i}]] + \Sigma[ \Ep[D_c(\hat x_t)\big|\hat x_t = \s_{i}] ] \stackrel{\mathclap{\normalfont\mbox{\footnotesize{\eqref{eq::cov_prop}}}}}{=}~  \Sigma[\hat x_{t+1}]
}
where the second expression in \eqref{eq::cov_computation} equals $\Sigma[ \Ep[D_c(x_t)] ]$ and third expression  is 0 because $\sum_j w_{i}^j \Ep[D_c(\s_{i})]=\Ep[D_c(\s_{i})] \left(\sum_j w_i^j\right) =  \Ep[D_c(\s_{i})]$ as $\sum_j w_i^j=1$ and therefore $\sum_j w_i^j \s_i^j=\Ep[D_c(\s_{i})]$.
\end{proof}

\subsection{Approximating chance constraints with robust constraints}
\label{sec:chance-constraint-approximation}
\label{appendix::b}
The analysis in this section is similar to ones presented in \cite{nemirovski2009, nemirovski2007}. 
In our case, the chance constraint \eqref{eq::chance constraint_GD} is of the form
\begin{align}
P\big[\underbrace{\begin{bmatrix}a^T & b\end{bmatrix}}_{\alpha^T}\begin{bmatrix}d\\1\end{bmatrix}\geq 0\big]\geq 1-\epsilon,
\label{eq::chance constraint}
\end{align}
where $a^T=\nabla_\theta c_i(x_{t+\tau})$, $b=c_i(x_{t+\tau})$, $\alpha=\begin{bmatrix}a^T\\b\end{bmatrix}\in\mathbb R^{\kappa+1}$. We assume that $\alpha$ can be written as a linear parametrization $$\alpha = \alpha_0+\sum_{l=1}^L\alpha_l\zeta_l,$$ where $\zeta_l$ are perturbation variables. Now, in the case that $\zeta_l$ are known to be independent, zero mean, random variables varying in segments $[-\sigma_l,\sigma_l]$, then \eqref{eq::chance constraint} can be safely approximated by the conic quadratic inequality
\begin{align}
w_0-\sqrt{2\ln\frac{1}{\epsilon}}\sqrt{\sum_{l=1}^L{\sigma_l}^2{w_l}^2}\geq 0.
\label{eq::approximate_chance_constraint}
\end{align}
where by definition $\alpha^T\begin{bmatrix}d\\1\end{bmatrix}$ is rewritten for compactness as:
\begin{align}
\left(\alpha_0+\sum_{l=1}^L \zeta_l \alpha_l\right)^T\begin{bmatrix}d\\1\end{bmatrix} \equiv w_0[d]+\sum_{l=1}^L \zeta_l w_l[d]\label{eq::alpha-w},
\end{align}
where $w_l[d]$, $l\in\{0,\dots,L\}$, are affine functions of $d$. Hence, obtaining $w_l$ provides a safe, approximate constraint (safe in the sense that the solution of the approximate problem is guaranteed to be a solution of the original problem as well). 
\\
To this end, we note that in our case we have:
\begin{align}
\alpha=\begin{bmatrix}
\nabla_\theta(c_i(x_{t+\tau})) \\
c_i(x_{t+\tau})
\end{bmatrix} = \begin{bmatrix}
\nabla_{\theta_1}(c_i(x_{t+\tau}))\\
\vdots  \\ 
\nabla_{\theta_\kappa}(c_i(x_{t+\tau})) \\
c_i(x_{t+\tau})
\end{bmatrix}
\end{align}
and that we can consider the Taylor series first-order approximation of each element of the vector $\alpha$ around the mean value $\Ep[x_{t+\tau}]$ of $x_{t+\tau}$, denoted for compactness $\hat x_{t+\tau}$. For example:
\begin{align}
c_i(x_{t+\tau})&=c_i(\hat x_{t+\tau})+\frac{\partial c_i(x_{t+\tau})}{\partial x_{t+\tau}}\big|_{\hat x_{t+\tau}}\tilde x_{{t+\tau}}%+\nonumber\\
%&+\frac{1}{2}\frac{\partial^2 c_i(x_{t+\tau})}{\partial x_{t+\tau}^2}\big|_{\hat x_{t+\tau}}(\tilde x_{{t+\tau}})^2+\dots,
\end{align}
where $\tilde x_{t+\tau} \triangleq x_{t+\tau} - \hat x_{t+\tau}$. For compactness, we drop the subscript $t+\tau$ and denote $\nabla_{\theta_\kappa}(c_i(\cdot))=g_\kappa(\cdot)$, so that 
\begin{align}
\alpha = \begin{bmatrix}g_1(x)\\\vdots\\g_\kappa(x)\\c_i(x)\end{bmatrix}=\begin{bmatrix}g_1(\hat x)+\frac{\partial g_1(x)}{\partial x}\big|_{\hat x}\tilde x\\ \vdots\\g_\kappa(\hat x)+\frac{\partial g_\kappa(x)}{\partial x}\big|_{\hat x}\tilde x\\
c_i(\hat x)+\frac{\partial c_i(x)}{\partial x}\big|_{\hat x}\tilde x\end{bmatrix},
\end{align}
which further yields 
\begin{align}
\alpha^T\begin{bmatrix}d\\1\end{bmatrix}&=c_i(\hat x)+\underbrace{\frac{\partial c_i(x)}{\partial x}\big|_{\hat x}}_{\left(\nabla_x c_i(x)\big|_{\hat x}\right)^T}\tilde x + \nonumber \\ &+ \left(\underbrace{g_1(\hat x)}_{\nabla_{\theta_1}c_i(\hat x)}+\underbrace{\frac{\partial g_1(x)}{\partial x}\big|_{\hat x}}_{(\nabla_x(\nabla_{\theta_1}c_i(x))\big|_{\hat x})^T}\tilde x\right)d_1 + \dots + \nonumber \\
&+ \left(\underbrace{g_\kappa(\hat x)}_{\nabla_{\theta_\kappa}c_i(\hat x)}+\underbrace{\frac{\partial g_\kappa(x)}{\partial x}\big|_{\hat x}}_{(\nabla_x(\nabla_{\theta_\kappa}c_i(x))\big|_{\hat x})^T}\tilde x \right)d_\kappa = \nonumber \\
&= c_i(\hat x) + (\nabla_x c_i(x)\big|_{\hat x})^T \tilde x + \nonumber \\&+d^T\nabla_\theta(c_i(\hat x)) + d^T\left( (\nabla_x(\nabla_\theta c_i(x))\big |_{\hat x})^T \tilde x \right),
\label{eq::it_was_a_torture}
\end{align}
where $w_0=c_i(\hat x)+d^T\nabla_\theta(c_i(\hat x))$, and the remaining terms correspond to the right-hand side of \eqref{eq::alpha-w} for $L=1$, with $\tilde x$ being a zero-mean vector (assuming an unbiased estimate $\hat x$ of $x$), which is confined within intervals $[-\sigma,\sigma]$ known by the quality of the estimation (e.g., using the covariance of the UT-based estimate in the earlier section). Hence, not surprisingly, what \eqref{eq::it_was_a_torture} and \eqref{eq::approximate_chance_constraint} reveal is that the constraint \eqref{eq::chance constraint_GD} can be approximated by a deterministic constraint that captures (and aims to counteract) the sources that contribute to possible violation of the constraint, namely the estimation error (via $\sigma$) and the behavior of the function around the current value via $\nabla_x(c_i(\cdot)), \nabla_x(\nabla_\theta(c_i(\cdot)))$. In practice it might be difficult to know bounds a priori for $\nabla_x(c_i(\cdot)), \nabla_x(\nabla_\theta(c_i(\cdot)))$ at each point $x\in \mathbb R^n$. However, local or global estimates of the Lipschitz constants of the corresponding functions might be available, and in that case they can provide a good estimate of the bounds. In summary, the chance constraint \eqref{eq::chance constraint_GD} can be approximated as:   
\begin{align}
\label{eq::approximate-constraint-appendix}
c_i(\Ep[x_{t+\tau}])+d^T\nabla_\theta(c_i(\Ep[x_{t+\tau}])+\sqrt{2\ln\frac{1}{\epsilon}}\Omega\geq 0,
\end{align}
where $\Omega=\omega(\Sigma[x_{t+\tau}], \nabla_x(c_i(\cdot)), \nabla_x(\nabla_\theta(c_i(\cdot))))$, $\omega:\mathbb R^n\rightarrow \mathbb R$ is a function that captures the statistics of the state, and the behavior of the constraint functions around the estimates of the state and the parameter.

\subsection{Approximating the chance constraint with an equivalent form}
\label{appendix::c}
We note that it is computationally inefficient for the purpose of online implementation to evaluate the gradients $\nabla_x(c_i(\cdot)), \nabla_x(\nabla_\theta(c_i(\cdot))))$ involved in \eqref{eq::approximate-constraint-appendix}. Therefore, in the implementation of case studies 3 and 4, which involve the tuning of the controller parameters, we consider the following constraint:
\eqn{
  &CF_i(x_{t+\tau}, \epsilon) + d^T \nabla_{\theta} ( CF_i(x_{t+\tau}, \epsilon) ) \geq 0,
  \label{eq::confidence_func_approx}
}
where the confidence function $CF:\reals^n \times [0,1]\rightarrow \reals$ is defined such that
\eqn{
CF_i(x,\epsilon)\geq 0 \implies \probs[c_i(x)\geq 0] \leq 1-\epsilon.
}
Eq. \eqref{eq::confidence_func_approx} could be an exact robust formulation of the chance constraint \eqref{eq::rl_constraint}, or a conservative approximation. For example, supposing $c(x)$ is given by a Gaussian distribution, one choice of $CF$ is as follows
\eqn{
  CF(x,\epsilon) = \Ep[c_i(x)] - \beta(\epsilon)\sqrt{\Sigma[c_i(x)]},
  \label{eq::gaussian_CF}
}
where $\beta:[0,1]\rightarrow \reals^+$ relates the probability $\epsilon$ to a factor of standard deviation that defines the confidence interval of Gaussian in \eqref{eq::gaussian_CF}. This factor is usually looked up in a table. For example, for $\epsilon=0.95$, $\beta(\epsilon)=1.96$. Eq. \eqref{eq::gaussian_CF} is based on the confidence interval of a Gaussian situated symmetrically about the mean, which is more conservative than a confidence interval that includes the tail and exactly solves \eqref{eq::rl_constraint}. Our simulation results are based on \eqref{eq::gaussian_CF} for the lack of analytical formulas for confidence intervals with higher order moments of a distribution.
\\

%we consider the terms that depend on the estimation error $\tilde x$ to be normally distributed with zero mean and known covariance $\Sigma[x_{t+\tau}]$, and approximate the constraint \eqref{eq::it_was_a_torture} as  
%\begin{align}
%\label{eq::approximate-constraint-case-studies}
%c_i(\Ep[x_{t+\tau}])+d^T\nabla_\theta(c_i(\Ep[x_{t+\tau}])+\beta(\epsilon)\sqrt{d^T\Sigma[x_{t+\tau}]d}\geq 0,
%\end{align}
%where $\beta(\epsilon)$ is 

\end{document}